# GContextFormer: A global context-aware hybrid multi-head attention approach with scaled additive aggregation for multimodal trajectory prediction

Yuzhi Chen[a], Yuanchang Xie[b], Lei Zhao[a], Pan Liu[c], Yajie Zou[d], Chen Wang[a,*]

[a] *Intelligent Transportation System Research Center, Southeast University, 2 Southeast University Road, Jiangning District, Nanjing, 211189, P.R. China*

[b] *Department of Civil and Environmental Engineering, University of Massachusetts Lowell, 1 University Ave, Lowell, MA 01854, USA*

[c] *Jiangsu Key Laboratory of Urban ITS, School of Transportation, Southeast University, Nanjing 211189, P.R. China*

[d] *Key Laboratory of Road and Traffic Engineering of Ministry of Education, Tongji University, Shanghai 201804, P.R. China*

## ABSTRACT

Multimodal trajectory prediction generates multiple plausible future trajectories to address vehicle motion uncertainty from surrounding-induced intention ambiguity and state-dependent execution variability. However, high-definition map–dependent models are limited by costly data acquisition, delayed updates, and vulnerability to stale or corrupted map inputs, causing brittle prediction failures in dynamic environments. In contrast, map-free approaches lack global context conditioning, with pairwise attention over-amplifying dominant straight patterns and suppressing minority transitional patterns, resulting in motion-intention misalignment with historical observations. This paper proposes GContextFormer, a plug-and-play encoder–decoder architecture with global context-aware hybrid multi-head attention and scaled additive aggregation that achieves intention-aligned multimodal prediction and balanced social interaction reasoning without map reliance. The Motion-Aware Encoder (MAE) builds a scene-level intention prior via bounded scaled additive aggregation over mode-embedded trajectory tokens and refines per-mode representations under a shared global context, mitigating inter-mode suppression and promoting intention alignment with historical observations. The Hierarchical Interaction Decoder (HID) decomposes social reasoning into dual-pathway cross-attention where a standard pathway ensures uniform geometric coverage over agent–mode pairs while a neighbor-context–enhanced pathway emphasizes salient interactions, with a gating module mediating their contributions to maintain coverage–focus balance. Experiments on eight highway-ramp scenarios from the TOD-VT dataset demonstrate that GContextFormer outperforms state-of-the-art baselines. Compared to existing transformer-based models, GContextFormer achieves greater robustness and concentrated improvements in high-curvature and transition zones via performance spatial distributions. Interpretability is achieved through motion mode distinctions and neighbor context modulation, which expose reasoning attribution. The modular architecture supports extensibility toward transportation and cross-domain multimodal reasoning tasks. The source code is available at https://fenghy-chen.github.io/sources/.

**Keywords:** Multimodal prediction, Context-aware, Scaled additive aggregation, Interaction reasoning, Uncertain motion

\* *Corresponding author.*

E-mail addresses: y.z.chen.fhy@gmail.com, yuzhi_chen@seu.edu.cn (Yuzhi Chen), yuanchang_xie@uml.edu (Yuanchang Xie), lei_zhao@seu.edu.cn (Lei Zhao), liupan@seu.edu.cn (Pan Liu), zouyajie@tongji.edu.cn (Yajie Zou), chen_david_wang@seu.edu.cn (Chen Wang).

**1 INTRODUCTION**

Vehicle trajectory prediction faces significant challenges from uncertain motion, which encompasses both diverse potential path selections based on driving intentions and stochastic variations in motion dynamics even when the intended path is predetermined (Huang et al., 2025; G. Li et al., 2024). This uncertainty becomes particularly pronounced in scenarios with multiple viable driving paths (e.g., unsignalized intersections) or expansive maneuverable spaces (e.g., unmarked/unstructured road segments, multi-lane dashed-line sections), such as highway-ramp merging or diverging zones where both categories of uncertainty converge and interact dynamically. Multimodal trajectory prediction (MTP) has attracted tremendous attention (Huang et al., 2025; Yang et al., 2025) due to its potential to capture and quantify this inherent motion uncertainty by inferring multiple plausible future trajectories along with their associated probability distributions, thereby enabling more robust and reliable downstream planning and risk-aware decision-making (Chen et al., 2025; Madjid et al., 2026; Zhao et al., 2025). Most existing models for vehicle trajectory prediction rely heavily on high-definition (HD) maps for contextual information, yet suffer from limited coverage and vulnerability to stale or adversarial errors that cause brittle prediction breakdowns, hindering their practical deployment and widespread adoption (Gu et al., 2024; He et al., 2025; Liao et al., 2024). Conversely, existing models without explicit map reliance lack global trajectory-mode context conditioning, resulting in motion-intention misalignment where predicted futures become inconsistent with historical observations. This paper aims to bridge this critical gap by proposing GContextFormer, a global context-aware encoder-decoder framework that achieves intention-aligned multimodal prediction without requiring HD map inputs.

The limitations of HD map-dependent models stem from multiple interconnected constraints. Limited coverage arises from the high cost of large-scale data acquisition, prohibitive expenses of map updates and maintenance, and inevitable obsolescence of map information in dynamic traffic environments (Gu et al., 2024; Liao et al., 2024). Even when coverage exists, stale or incomplete updates in evolving environments (e.g., temporary lane closures, construction detours, parked-vehicle churn) can inject erroneous geometric constraints or navigation cues, while the adversarial or corrupted map inputs can systematically bias the scene understanding. Both can precipitate brittle breakdowns in prediction when models over-rely on map cues, introducing substantial risks that can cascade into collision scenarios in critical applications. Existing MTP models predominantly leverage HD maps through three main paradigms. Rasterized representations convert vectorized map data into multi-channel bird's eye view (BEV) images, which are subsequently processed by convolutional neural networks to extract spatial-semantic features from the discretized map information (L. Li et al., 2024; Liu et al., 2023; Xu et al., 2023; Zhang et al., 2024). Vectorized representations directly encode raw polyline information from HD maps, preserving geometric precision while enabling efficient processing through transformer architectures that can handle sequential map elements (Chen et al., 2025; Liao et al., 2025; Tang et al., 2024). Graph-structured representations model lane topology as graph networks with nodes representing lane segments and edges capturing spatial-temporal relationships, processed via graph neural networks for context-aware feature aggregation (Carrasco Limeros et al., 2023; Deo et al., 2022; Wang & Sun et al., 2024). These approaches integrate map information with historical trajectory data through various fusion mechanisms, including early fusion at the feature extraction stage, intermediate fusion via attention mechanisms during encoding, and late fusion through constraint optimization during prediction, to generate contextually informed multimodal trajectory forecasts (Chen et al., 2025; Huang et al., 2023; Liu et al., 2023; Tang et al., 2024).

Recognizing the limitations of HD map-dependent MTP modeling, recent research has increasingly advocated for lightweight and map-free MTP designs to meet broader application demands (Kawasaki & Seki, 2021; Sang et al., 2026), particularly for scenarios such as vehicle collision risk modeling in newly constructed or reconstructed roads and autonomous driving motion planning in unmapped environments. These emerging approaches demonstrate promising potential by relying primarily on historical trajectory patterns and inter-agent interactions to capture contextual information (Liao et al., 2024; Liu et al., 2024; Xiang et al., 2024). Behavior-driven approaches that construct dynamic geometric graphs to capture continuous driving behaviors and social interactions, such as MFTraj which assumes that neighboring vehicles' historical movements implicitly encode road structure information (Liao et al., 2024), and

*Chen et al.*Social-STGCNN (Mohamed et al., 2020) or Social-Implicit (Mohamed et al., 2022), which further enhance spatio-temporal relation modeling through graph convolution and interaction-latent optimization to better approximate continuous motion patterns. Multi-level spatial-temporal modeling approaches employ hierarchical attention mechanisms to extract context from historical trajectory sequences at different temporal scales, attempting to compensate for the absence of explicit road geometry through temporal pattern learning (Lei et al., 2025; Wu et al., 2021; Xiang et al., 2024). Such models include STGAT (Huang et al., 2019) which introduces dual-pathway attention to model both spatial interactions via graph attention networks and temporal correlations of interactions through dedicated LSTM modules, and TUTR (Shi et al., 2023), a unified transformer framework that eliminates post-processing requirements through motion mode parsing and social-level attention decoding. Social interaction networks focus on modeling inter-vehicle relationships through graph neural networks (Lu et al., 2025; Xiang et al., 2024) or attention mechanisms (Peng et al., 2022), treating trajectory prediction as a collective behavior modeling problem where individual vehicle movements are influenced by surrounding traffic dynamics (Cheng et al., 2023; Si et al., 2024). For instance, Multiclass-SGCN (Li et al., 2022) which addresses multi-class trajectory prediction via sparse graph convolution with adaptive interaction masking to accommodate heterogeneous agent categories and improve prediction robustness under varying densities.

However, the above map-free models are limited by inadequate context integration arcoss potential motion patterns (Madjid et al., 2026). Existing pairwise attention mechanisms (e.g., transformer-based models) over tokens fail to establish global context of trajectory-mode associations, leading to intention misalignment (decision-point divergence from the historical motion trend) and shape inconsistency (curvature/shape deviation from the ground-truth future) (Shi et al., 2023; Wang & Zhang et al., 2024; Yang et al., 2024). In high-curvature ramp scenarios, dominant straight-trajectory patterns may suppress equally plausible turning modes, yielding under-steering predictions that cluster around prevalent behaviors while missing the intended curved maneuver. Also, dominant motion patterns may overshadow equally plausible alternatives, resulting in conservative predictions that cluster around prevalent behaviors rather than capturing the likelihood spectrum of uncertain motion. Additionally, these approaches find it difficult to effectively balance target vehicle individual behaviors with collective neighborhood dynamics. Current methods model pair-wise interactions without considering the hierarchical nature of individual versus collective motion patterns, where such imbalanced social interactions disrupt the cognitive hierarchy of motion decision-making (Shi et al., 2023; Xu et al., 2022; Zheng et al., 2021). This potentially leads to biased representation of multi-level social dependencies and unstable prediction accuracy across various scenarios. In mainline-ramp transition zones, over-emphasizing immediate neighbor interactions while neglecting collective motion trends context can result in trajectory predictions that fail to distinguish between lane-keeping and lane-changing intentions, compromising decision-point accuracy.

To address these limitations, this paper proposes GContextFormer, a plug-and-play encoder-decoder architecture with global context-aware hybrid multi-head attention and scaled additive aggregation, achieving intention-aligned multimodal prediction and balanced social interaction reasoning without reliance on map inputs. The Motion-Aware Encoder (MAE) constructs a scene-level intention prior via bounded scaled additive aggregation over mode-embedded trajectory tokens and refines per-mode representations under a shared global context while preserving inter-mode distinctions. This mitigates inter-mode suppression, promoting intention alignment and shape consistency with historical observations. This design enables collaborative refinement where each mode-embedded trajectory token benefits from shared global context while maintaining distinctive behavioral characteristics. The Hierarchical Interaction Decoder (HID) decomposes social reasoning into a parameter-shared dual-pathway architecture where a standard pathway ensures uniform geometric coverage over agent-mode pairs and a neighbor-context-enhanced pathway emphasizes salient interactions under neighbor context prior, with a gating module mediating their contributions to maintain a coverage-focus balance. This hierarchical design addresses the challenge of balancing individual target vehicle behaviors with collective neighborhood dynamics through learnable attention allocation. The effectiveness of the proposed modular framework is demonstrated using extensive trajectory data collected from eight highway-ramp scenarios, where the complementary encoder-decoder design enables robust multimodal prediction across diverse geometric configurations and interaction complexities.

## 2 GCONTEXTFORMER ARCHITECTURE

Our model aims to forecast multiple plausible future trajectories for a target vehicle without relying on HD map information. This motivation leads to the design of context-modeling strategies that remain effective even under partial, coarse, or unreliable map conditions. The model primarily utilizes the historical trajectories of the target vehicle and its surrounding agents to generate diverse future motion hypotheses, thereby capturing both intention ambiguity and execution variability inherent in human driving behavior. As illustrated in **Figure 1**, the proposed GContextFormer follows a modular encoder–decoder architecture that integrates global motion understanding with hierarchical social reasoning. The architecture begins with the general motion modes estimation module, which processes normalized historical trajectories to produce motion-mode representations. These representations serve as the encoded inputs to the motion-aware encoder (MAE), where local trajectory features and global contextual cues are fused to construct motion-conditioned embeddings. The resulting latent features are then passed to the hierarchical interaction decoder (HID), which progressively refines them through multi-agent social reasoning to infer coherent, multimodal future trajectories. The overall pipeline captures both temporal dependencies and interaction dynamics, leading to intention-aligned multimodal trajectory predictions with balanced individual and collective behavior modeling.

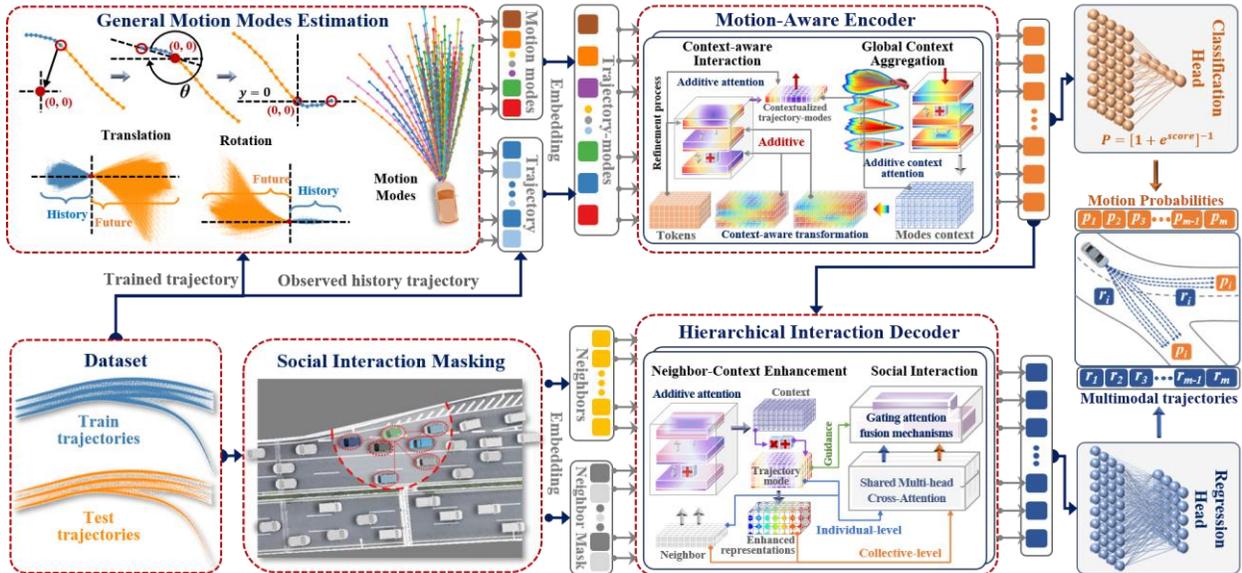

**Figure 1 GContextFormer model architecture**

### 2.1 Definition and Formulation

*2.1.1 Problem definition*

Our model aims to predict multiple plausible future trajectories and their probabilities for a target vehicle without relying on HD map information, using only observed trajectories of the target vehicle and its neighboring vehicles. Consider a traffic scene containing N vehicles, where vehicle $i = 0$ represents the target vehicle and vehicles $i = 1, 2, \ldots, N-1$ denote neighboring vehicles. For each vehicle i, we observe its historical trajectory over $T_{obs}$ observation time steps, denoted as $X_i^{obs} = \left(x_i^t, y_i^t\right)_{t=1}^{T_{obs}}$, where $(x_i^t, y_i^t)$ represents the 2D coordinates at time step t. The target vehicle's historical trajectory is $X_0^{obs} = \{(x_0^t, y_0^t)\}_{t=1}^{T_{obs}}$, while the neighboring vehicles' historical trajectories form the set $X_{neig}^{obs} = \{X_i^{obs}\}_{i=1}^{N-1}$.

Given these historical observations, our objective is to predict $K$ diverse future trajectories for the target vehicle over $T_{pre}$ prediction time steps, formulated as $\hat{Y}_0 = \{\hat{Y}_0^k, p_k\}_{k=1}^K$, where each predicted trajectory $\hat{Y}_0^k = \{(\hat{x}_0^t, \hat{y}_0^t)\}_{t=T_{obs}+1}^{T_{obs}+T_{pre}}$ is associated with probability $p_k$ satisfying $\sum_{k=1}^K p_k = 1$. The ground truth future trajectory of the target vehicle is defined as $Y_0^{gt} = \{(x_0^{t,gt}, y_0^{t,gt})\}_{t=T_{obs}+1}^{T_{obs}+T_{pre}}$.

*2.1.2 Trajectories normalization*

Given our model operates without HD map information, it must be invariant to absolute positions and orientations to focus on relative motion patterns rather than specific coordinate values. Without proper normalization, models may overfit to specific absolute positions and orientations present in the training data, limiting their ability to generalize across different geographic locations and road orientations. To address this challenge and enhance the model's robustness to coordinate system variations, we apply translation and rotation operations to normalize all trajectories. Translation shifts the entire scene so that the target vehicle's final observed position $(x_0^{T_{obs}}, y_0^{T_{obs}})$ becomes the new coordinate origin. Specifically, each coordinate $(x_i^t, y_i^t)$ is transformed to $(x_i^t - x_0^{T_{obs}}, y_i^t - y_0^{T_{obs}})$. Subsequently, rotation normalizes the spatial orientation by aligning the target vehicle's initial position vector in the translated coordinate system with a reference direction. The translated first observation position is $(x_0^1 - x_0^{T_{obs}}, y_0^1 - y_0^{T_{obs}})$, and its direction angle is computed as $\theta = atan2(y_0^1 - y_0^{T_{obs}}, x_0^1 - x_0^{T_{obs}})$, where $atan2(y, x)$ denotes the four-quadrant arctangent function that computes the angle in radians between the positive x-axis and the point $(x, y)$. We then apply the rotation matrix $R = \begin{bmatrix} \cos\theta & -\sin\theta \\ \sin\theta & \cos\theta \end{bmatrix}$ to each trajectory point, ensuring the transformed target vehicle's initial position aligns with the positive x-axis. This coordinate normalization enables the model to learn relative motion patterns that are invariant to absolute position and orientation variations.

*2.1.3 General motion modes*

Given the multimodal nature of trajectory prediction where a single ground truth trajectory cannot capture all reasonable future possibilities, a set of representative motion patterns is essential to guide the generation of multiple trajectory candidates. These patterns, termed general motion modes, are derived from normalized trajectories that enable comparable representation of behavioral patterns across different scenarios. Specifically, normalized trajectories allow common relative motion behaviors, such as car-following, lane-changing, going straight, turning left/right, and yielding maneuvers, to be compared and clustered effectively regardless of their original absolute geographic positions or road orientations. General motion modes provide structural templates that reflect fundamental driving behaviors, serving as prior knowledge to guide the model's trajectory generation process.

To extract general motion modes, we apply k-means clustering to the normalized ground truth trajectories from the training dataset. Let $\widetilde{Y}_{gt} = \{\tilde{Y}_j^{gt}\}_{j=1}^M$ denote the collection of M normalized ground truth future trajectories, where each trajectory $\tilde{Y}_j^{gt} = \{(\tilde{x}_j^t, \tilde{y}_j^t)\}_{t=T_{obs}+1}^{T_{obs}+T_{pre}}$ represents the normalized future path of the j-th target vehicle. Each trajectory is flattened into a feature vector $\boldsymbol{f}_j \in \mathbb{R}^{2T_{pre}}$ by concatenating all coordinate pairs, $\boldsymbol{f}_j = [\tilde{x}_j^{T_{obs}+1}, \tilde{y}_j^{T_{obs}+1}, \ldots, \tilde{x}_j^{T_{obs}+T_{pre}}, \tilde{y}_j^{T_{obs}+T_{pre}}]^T$. The k-means algorithm partitions these trajectory features into K clusters, where K corresponds to the desired number of motion modes. The clustering objective minimizes the within-cluster sum of squares:

$$\arg\min_{C,A} \sum_{j=1}^M \sum_{k=1}^K a_{jk} \|\boldsymbol{f}_j - \boldsymbol{c}_k\|_2^2 \qquad (1)$$

where $\boldsymbol{C} = \{\boldsymbol{c}_k\}_{k=1}^K$ represents the set of cluster centroids, $\boldsymbol{A} = \{a_{jk}\}$ denotes the assignment matrix with $a_{jk} = 1$ if trajectory j belongs to cluster $k$ and $a_{jk} = 0$ otherwise, and $\|\cdot\|_2$ represents the Euclidean norm.

Upon convergence, each cluster centroid $c_k$ is reshaped back to trajectory format to form the k-th general motion mode, $M_k^{motion} = \{(c_k^{t,x}, c_k^{t,y})\}_{t=T_{obs}+1}^{T_{obs}+T_{pre}}$, where $(c_k^{t,x}, c_k^{t,y})$ represents the coordinates at time step t for motion mode k. The complete set of general motion modes is defined as $\mathcal{M} = \{M_k^{motion}\}_{k=1}^{K}$. These motion modes capture the most representative trajectory patterns in the training data and serve as structural priors for generating diverse and realistic trajectory predictions during inference.

*2.1.4 Social interaction masking*

While neighboring vehicles significantly influence target vehicle behavior, not all vehicles in a traffic scene contribute equally to the prediction task, and scenarios contain varying numbers of relevant neighbors. To identify spatially relevant neighbors and maintain unified tensor dimensions across scenes, a preliminary social interaction masking mechanism is adopted that performs coarse spatial filtering before the detailed interaction reasoning conducted in the Hierarchical Interaction Decoder (HID). For each neighbor $i$, the average Euclidean distance to the target vehicle over the observation period is calculated (prior to normalization) as

$$d_i = \frac{1}{T_{obs}} \sum_{t=1}^{T_{obs}} \left\| (x_i^t, y_i^t) - (x_0^t, y_0^t) \right\|_2 \tag{2}$$

Neighbors that remain within a predefined distance threshold are considered effective participants and retained in the interaction set $\mathcal{N}^{eff} = \{i \in \{1,2,\ldots,N-1\}: d_i < \delta\}$, where $\delta$ is a predefined threshold. Let $n_j = \left|\mathcal{N}_j^{eff}\right|$ be the number of effective neighbors for the $j$-th scene in a batch, and let $N_{max}^{neig} = \max_j\{n_j\}$ the maximum neighbors count among all scenes. To ensure tensor consistency, zero-padding is applied to scenes with fewer than $N_{max}^{neig}$ neighbors. The social interaction mask is formalized as a binary matrix $M_j^{neig} \in \{0,1\}^{N_{max} \times N_{max}}$ for each scene $j$, defined as

$$M_j^{neig}[i,k] = \begin{cases} 1, & \text{if } i < n_j \text{ and } k < n_j \\ 0, & \text{otherwise} \end{cases} \tag{3}$$

This mask distinguishes actual neighbors from padded entries during attention computation, preventing spurious patterns from dummy neighbors and supporting pairwise interaction modeling. The neighbor trajectory tensor has the dimension $[B, N_{max}^{neig}, T_{obs}, 2]$, and the corresponding mask tensor is $[B, N_{max}^{neig}, N_{max}^{neig}]$. Together, they provide a structured representation of the local neighborhood that enables batch-level efficiency and consistent pairwise processing. The distance metric in this masking mechanism offers a coarse but efficient approximation of spatial relevance rather than an exact measure of interactive importance. In practice, this preliminary filtering ensures that only neighbors with sustained spatial proximity are carried forward for higher-level reasoning. Dynamic or asymmetric interaction effects, such as overtaking, lane merging, or yielding, are not directly modeled at this stage. These finer and directional relations are explicitly captured by the Hierarchical Interaction Decoder (HID), where attention mechanisms learn temporally adaptive and direction-sensitive dependencies among interacting vehicles.

**2.2 Motion-Aware Encoder (MAE)**

The motion-aware encoder (MAE) constitutes the core component for processing trajectory-mode associations and extracting contextualized motion representations. Unlike conventional encoder architectures that treat multiple trajectory hypotheses independently, this encoder introduces a collaborative refinement mechanism that enables each mode-embedded trajectory token to benefit from shared global context while preserving its distinctive behavioral characteristics. Instead of relying on pairwise token interactions alone, MAE builds a global motion context that functions as a scene-level intention prior (global context $G$) and conditions subsequent per-mode updates. As illustrated in **Figure 2**, scaled additive attention is adopted for global context aggregation and context-aware transformation, with separate projections for global and per-mode terms. This parameterization clarifies the roles of global and local signals within the scoring function and emphasizes learned correspondence in the projected space. This conditioning is intended to mitigate inter-mode suppression and homogenization by balancing shared priors with per-mode distinctions.

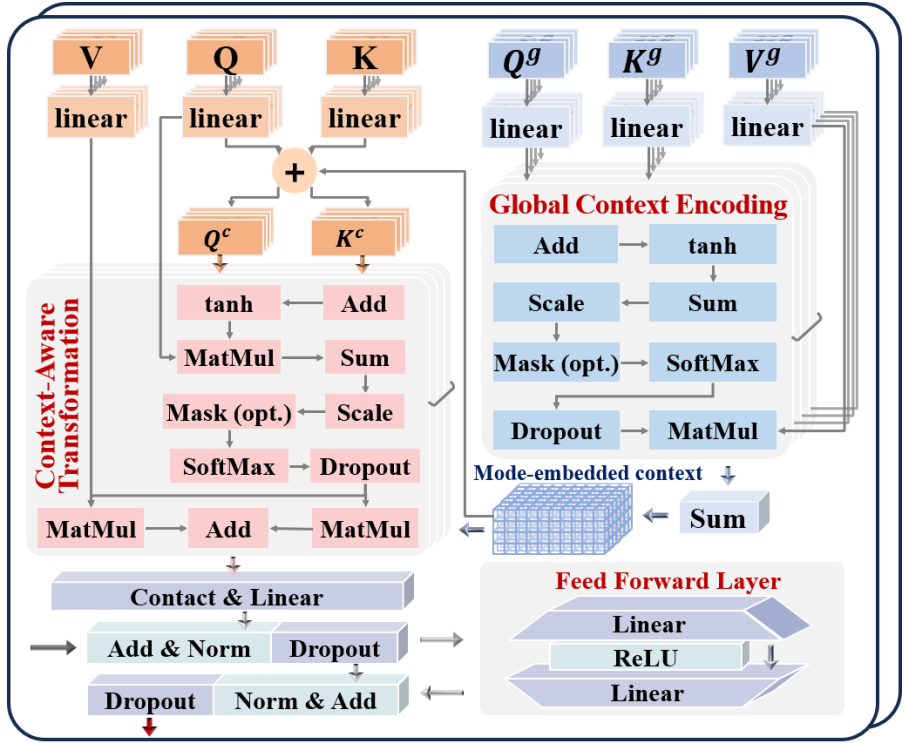

**Figure 2 Motion-aware encoder (MAE) architecture.**

*2.2.1 Global context aggregation*

For a predicted scene, the input consists of the standardized observed trajectory of the target vehicle $\tilde{X}_i^{obs}$ and the set of $K$ general motion modes $\boldsymbol{M} = \{\boldsymbol{M}_k^{motion}\}_{k=1}^{K}$. For each motion mode $k$, a mode-embedded trajectory token is constructed in Eq. (4) by temporal concatenation of the observed trajectory and the $k$-th motion mode.

$$\boldsymbol{S}_k = \left[\tilde{X}_i^{obs}; \boldsymbol{M}_k^{motion}\right] \in \mathbb{R}^{(T_{obs}+T_{pre}) \times 2} \tag{4}$$

where [ ; ] denotes temporal concatenation. This concatenation enables the model to learn how historical motion patterns relate to potential future behaviors encoded in each motion mode, which supports subsequent feature embedding and context aggregation. Each mode-embedded trajectory token $\boldsymbol{S}_k$ is then flattened and mapped via a fully connected layer to an embedding vector $\boldsymbol{E}_k$ in Eq. (5).

$$\boldsymbol{E}_k = \text{Linear}_{\text{embedding}}\big(\text{flatten}(\boldsymbol{S}_k)\big) \in \mathbb{R}^{d_{model}} \tag{5}$$

where $d_{model}$ is the embedding dimension and the flatten operation converts the 2D trajectory sequence into a 1D feature vector of length $2(T_{obs} + T_{pre})$. The resulting set $\{\boldsymbol{E}_k\}_{k=1}^{K}$ forms the mode-token sequence for further context encoding.

Global aggregation is performed via scaled additive attention, which applies a Bahdanau-style additive attention (Bahdanau et al., 2014) with a scaling factor $1/\sqrt{d_k}$ before softmax(Vaswani et al., 2017), to capture scene-level commonalities across candidate modes while prioritizing semantic alignment in a learned projection space. In this formulation, a global context vector summarizes shared tendencies across modes. The scoring function uses separate projections for the global and per-mode inputs, which delineates

*Chen et al.*

their roles and reduces reliance on raw feature magnitudes or incidental alignments. The global context $\boldsymbol{G} \in \mathbb{R}^{d_{model}}$ is computed through weighted aggregation in Eq. (6).

$$\boldsymbol{G} = \sum_{k=1}^{K} \alpha_k \boldsymbol{v}_k^g \tag{6}$$

where $\boldsymbol{v}_k^g$ is the global value projection of embedding $\boldsymbol{E}_k$, and the attention weights $\alpha_k$ are derived from scaled additive attention scores computed using global query and key projections as expressed in Eq. (7).

$$\alpha_k = \mathrm{softmax}\left[\frac{\sum_{j=1}^{d_k} \tanh(\boldsymbol{q}_k^g + \boldsymbol{k}_k^g)}{\sqrt{d_k}}\right], \mathrm{softmax}(z_i) = \frac{\exp(z_i)}{\sum_{j=1}^{d_k} \exp(z_j)} \tag{7}$$

where $\boldsymbol{q}_k^g$ and $\boldsymbol{k}_k^g$ are the global query and key projections of $\boldsymbol{E}_k$, $d_k = d_{model}/h$ with $h$ being the number of attention heads, and the sum operates over the vector dimension. The softmax(·) function normalizes the input scores $z_i$ into a probability distribution, ensuring that all output values are non-negative and sum to unity, where $z_i$ represents the $i$-th element of the input vector. Within the additive scoring, the bounded nonlinearity tanh(·) yields smoothly saturating responses that stabilize aggregation, which favors a coherent global summary across modes and reduces undue influence from high-magnitude or incidentally aligned inputs. The normalization by $\sqrt{d_k}$ controls score scale and contributes to stable optimization. The resulting $\boldsymbol{G}$ serves as a compact scene-level context that summarizes motion tendencies shared across modes.

*2.2.2 Context-aware transformation attention*

Each trajectory-mode embedding interacts with the computed global context to produce refined representations that encode both local motion patterns and scene-level context. This transformation mitigates the pairwise-only limitation of standard attention by conditioning each embedding $\boldsymbol{E}_k$ on a shared reference. The global context $\boldsymbol{G}$ is used to nonlinearly modulate and align local features so that mode-specific cues are interpreted under a shared intention prior. The additive integration keeps global semantics consistent across all trajectory modes, ensuring each mode receives identical global semantic guidance for fair comparison while preserving their distinctive local characteristics. The context-aware transformation enhances the standard query and key representations by incorporating global information using Eq. (8).

$$\boldsymbol{q}_k^{ctx} = \boldsymbol{q}_k + \boldsymbol{G}, \qquad \boldsymbol{k}_k^{ctx} = \boldsymbol{k}_k + \boldsymbol{G} \tag{8}$$

where $\boldsymbol{q}_k$ and $\boldsymbol{k}_k$ are the standard query and key projections of $\boldsymbol{E}_k$, respectively. This additive injection of $\boldsymbol{G}$ produces an intention-aligned shift of the matching space, allowing each mode to be refined by the same global semantic frame while retaining its distinctive evidence.

The attention scores $s_k^{en}$ for each mode are computed in Eq. (9) using the enhanced representations and the original query projection to maintain mode-specific characteristics. This mechanism enables each motion mode to dynamically adjust its representation informed by both its original context and the collective behaviors inferred from all candidate modes of the current scene. The final trajectory-mode representations $\boldsymbol{c}_k^{en}$ incorporate both attended local values and global context contributions through a collaborative refinement process in Eq.(10), where $\boldsymbol{v}_k$ represents the value projection of $\boldsymbol{E}_k$.

$$s_k^{en} = \frac{\sum_{j=1}^{d_k}[\tanh(\boldsymbol{q}_k^{ctx} + \boldsymbol{k}_k^{ctx}) \cdot \boldsymbol{q}_k]}{\sqrt{d_k}} \tag{9}$$

$$\boldsymbol{c}_k^{en} = \mathrm{softmax}(s_k^{en}) \cdot \boldsymbol{v}_k + \mathrm{softmax}(s_k^{en}) \cdot \boldsymbol{G} \tag{10}$$

The motion-aware encoder employs a multi-head structure where each head operates on a subspace of dimension $d_k = d_{model}/h$. The outputs from all heads are concatenated along the feature dimension and then linearly projected back to the full embedding dimension $d_{model}$, following the standard transformer multi-head attention formulation. The parallel use of a mode-specific value pathway and a global-context pathway is designed to maintain per-mode distinctions while introducing a shared scene prior. This structure is intended to reduce the risk of representation homogenization across modes. The output of the motion-aware encoder consists of the set $\boldsymbol{C}_{en} = \{\boldsymbol{c}_k^{en}\}_{k=1}^{K}$ and the corresponding score set $S_{en} = \{s_k^{en}\}_{k=1}^{K}$, where the scores quantify the consistency between the driving history and each motion hypothesis within the shared global context. These features provide both diversity and plausibility priors, serving as the foundation for subsequent trajectory scoring and social interaction modeling phases.

### 2.3 Hierarchical Interaction Decoder (HID)

The hierarchical interaction decoder (HID) refines contextualized trajectory-mode representations produced by the MAE by integrating social interaction dynamics from neighboring vehicles. As shown in Figure 3, HID employs a hierarchical attention architecture that decomposes social reasoning into a dual-pathway cross-attention mechanism. The neighbor-context-enhanced cross-attention pathway (green arrows) constructs a global neighbor context that captures collective spatial patterns across the neighborhood, while the standard cross-attention pathway (orange arrows) ensures uniform geometric coverage through direct pairwise interactions between trajectory-modes and individual neighbors. The decoder further introduces a learnable gating fusion to adaptively weight individual neighbor influences against collective neighborhood context, enabling robust trajectory prediction under diverse social interaction scenarios.

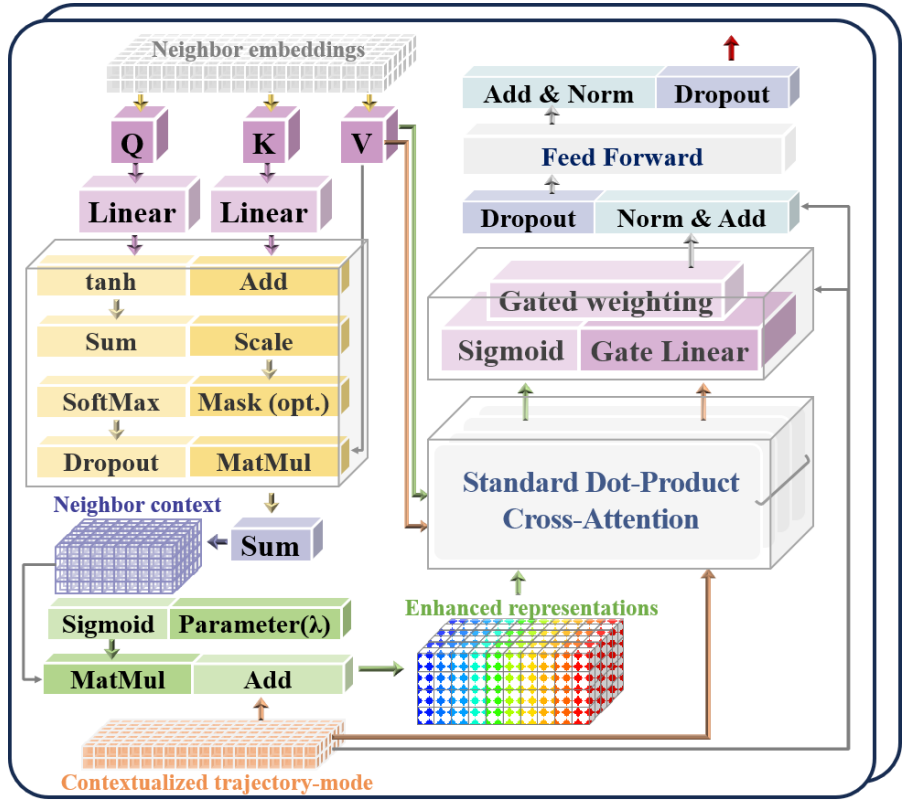

**Figure 3** Hierarchical interaction decoder (HID) architecture

#### 2.3.1 Neighbor-context-enhanced cross-attention

The decoder processes contextualized trajectory-mode features $\boldsymbol{C}_{en} = \{\boldsymbol{c}_k^{en}\}_{k=1}^{K}$ from the motion-aware encoder alongside neighboring vehicle information. Given the normalized neighboring trajectories $\widetilde{\boldsymbol{X}}_{neig}^{obs} = \{\widetilde{\boldsymbol{X}}_i^{obs}\}_{i=1}^{N_{max}^{neig}}$ and the social interaction mask $\boldsymbol{M}^{neig}$ defined, neighbor embeddings are computed in Eq. (11) through linear transformation.

*Chen et al.*

$$\mathbf{N}_i^{neig} = \text{Linear}_{\text{nei\_embedding}}\left(\text{flatten}\left(\widetilde{\mathbf{X}}_i^{obs}\right)\right) \in \mathbb{R}^{d_{model}} \quad (11)$$

where each neighbor trajectory is flattened into a feature vector of length $2T_{obs}$ before embedding. The resulting neighbor embeddings $\mathbf{N}^{neig} = \{\mathbf{N}_i^{neig}\}_{i=1}^{N_{max}^{neig}}$ form the contextual information for social interaction modeling.

The hierarchical interaction mechanism operates through two complementary attention pathways. The standard cross-attention pathway employs conventional dot-product attention to model direct pairwise interactions between each trajectory-mode and individual neighbors. For trajectory-mode k, the standard cross-attention output is computed as shown in Eq. (12).

$$\mathbf{O}_k^{std} = \text{CrossAttention}\left(\mathbf{c}_k^{en}, \mathbf{N}^{neig}, \mathbf{M}^{neig}\right) \quad (12)$$

where $\mathbf{c}_k^{en}$ serves as the query, $\mathbf{N}^{neig}$ provide keys and values, and $\mathbf{M}^{neig}$ masks out invalid or distant neighbors to restrict attention to effective ones. This pathway captures explicit neighbor-specific influences and maintains the capability to distinguish individual neighbor contributions.

Simultaneously, the neighbor-context-enhanced pathway computes a global neighbor context to capture collective spatial patterns that may not be evident in pairwise interactions. The global neighbor context $\mathbf{G}^{neig} \in \mathbb{R}^{d_{model}}$ is aggregated over valid neighbors using scaled additive attention with bounded scoring, which is derived in Eq. (13), where the attention weights $\beta_i$ are computed using scaled additive attention scores in Eq. (14). with $\mathbf{q}_i^{neig}$ and $\mathbf{k}_i^{neig}$ representing query and key projections of neighbor embedding $\mathbf{N}_i^{neig}$, respectively. The summation in Eq. (14) operates over the feature dimension, and $d_k = d_{model}/h$ with $h$ denoting the number of heads. The social interaction mask ensures that only effective neighbors contribute to the global context computation.

$$\mathbf{G}^{neig} = \sum_{i=1}^{N_{max}^{neig}} \beta_i \mathbf{N}_i^{neig} \quad (13)$$

$$\beta_i = \text{softmax}\left(\frac{\sum_{j=1}^{d_k} \tanh\left(\mathbf{q}_i^{neig} + \mathbf{k}_i^{neig}\right)}{\sqrt{d_k}}\right) \quad (14)$$

The neighbor-context-enhanced cross-attention pathway incorporates this global neighbor context into trajectory-mode representations before performing cross-attention. Each trajectory-mode embedding is enhanced through additive combination with the global neighbor context in Eq. (15), with a learnable fusion weight λ constrained to the range $[\lambda_{min}, \lambda_{max}]$ to prevent over-influence.

$$\mathbf{c}_k^{enh} = \mathbf{c}_k^{en} + \lambda \mathbf{G}^{neig} \quad (15)$$

where $\lambda = \lambda_{min} + (\lambda_{max} - \lambda_{min}) \cdot sigmoid(\lambda_{raw})$ with $\lambda_{\text{raw}}$ being a learnable parameter. The enhanced trajectory-mode representations then undergo cross-attention with neighbor information as shown in Eq. (16).

$$\mathbf{O}_k^{enh} = \text{CrossAttention}\left(\mathbf{c}_k^{enh}, \mathbf{N}^{neig}, \mathbf{M}^{neig}\right) \quad (16)$$

This enhancement mechanism enables each trajectory-mode to leverage both its original motion characteristics and the collective neighborhood context when establishing social interactions, facilitating a more comprehensive modeling of spatial neighborhood patterns.

*2.3.2 Gating attention fusion mechanisms*

The hierarchical decoder employs learnable gating mechanisms to adaptively balance the contributions from standard cross-attention and neighbor-context-enhanced cross-attention pathways. This fusion approach addresses the challenge of optimally combining individual neighbor influences with collective neighborhood context under varying traffic scenarios.

For each trajectory-mode k, a mode-specific gating weight $g_k$ is computed in Eq. (17) through a learnable transformation of the original trajectory-mode feature. The gating weight determines the relative importance of standard versus enhanced cross-attention outputs. When $g_k$ approaches 1, the model emphasizes direct pairwise neighbor interactions captured by standard cross-attention. Conversely, when $g_k$ approaches 0, the model prioritizes collective spatial patterns encoded through neighbor-context-enhanced attention. The input to the gating operation in Eq. (17) is regulated through linear weight normalization and bounded feature scaling inherited from the preceding layer normalization. The final socially-conditioned trajectory-mode representation then combines both attention pathways through weighted fusion as shown in Eq. (18).

$$g_k = \text{sigmoid}\left(\text{Linear}_{\text{gate}}(c_k^{en})\right) \in [0,1] \tag{17}$$

$$\boldsymbol{c}_k^{dec} = g_k \boldsymbol{O}_k^{std} + (1 - g_k)\boldsymbol{O}_k^{enh} \tag{18}$$

This gating mechanism enables the model to automatically adapt to different trajectory-mode characteristics and social interaction requirements. When a trajectory-mode benefits more from direct pairwise neighbor relationships, the learned gating weight $g_k$ increases, emphasizing standard cross-attention that captures explicit individual neighbor influences. Conversely, when a trajectory-mode requires broader spatial awareness, $g_k$ decreases, prioritizing neighbor-context-enhanced attention that incorporates collective spatial patterns from the global neighbor context. This adaptive balance is learned end-to-end, enabling the model to dynamically choose the optimal mix of individual neighbor influences and aggregate neighborhood context for each motion mode. To stabilize training and preserve mode-specific evidence, the decoder applies residual connections and layer normalization as in Eq. (19), ensuring that social interaction information augments rather than replaces the original motion-mode characteristics.

$$\boldsymbol{c}_k^{out} = \text{LayerNorm}\left(\boldsymbol{c}_k^{dec} + \boldsymbol{c}_k^{en}\right) \tag{19}$$

The decoder output consists of socially-conditioned trajectory-mode features $\boldsymbol{C}_{dec} = \{\boldsymbol{c}_k^{out}\}_{k=1}^{K}$ that integrate motion-aware encoding, individual neighbor interactions, and collective neighborhood context. Both cross-attention pathways employ multi-head attention, where each head operates on a subspace of dimension $d_k = d_{model}/h$ to capture diverse interaction cues. The head outputs are concatenated and linearly projected to restore the full embedding dimension, enabling the decoder to effectively combine fine-grained neighbor influences with broader contextual dependencies. Multiple HID layers can be stacked to capture increasingly complex social interaction patterns, with each layer reusing $\boldsymbol{M}^{neig}$ for masking and recomputing $\boldsymbol{G}^{neig}$ to reflect updated neighborhood context.

**2.4 Dual-Head Prediction**

The proposed model adopts a dual-head prediction framework that consists of a mode classification head and a trajectory regression head. This design enables the model to flexibly generate multiple socially-aware trajectory hypotheses with mode probabilities.

*2.4.1 Multi-trajectories and probabilities generator*

(1) Classification Head for Mode Selection

The classification head operates directly on encoded trajectory-mode representations to evaluate the likelihood of each motion pattern. Given the trajectory-mode features $\boldsymbol{C}_{en} = \{\boldsymbol{c}_k^{en}\}_{k=1}^{K}$ from the motion-aware encoder, the classification head computes confidence scores through linear transformation using Eq. (20).

$$\boldsymbol{s}_k^{cls} = Linear_{cls}(\boldsymbol{c}_k^{en}) \tag{20}$$

These scores $\boldsymbol{S}_{cls} = \{\boldsymbol{s}_k^{cls}\}_{k=1}^{K}$ serve dual purposes depending on the operational mode. During training, all $K$ scores are directly output for loss computation, enabling the model to learn comprehensive mode relationships across the entire trajectory-mode bank. During inference, the scores serve as criteria to select top-k likely motion modes and are transformed into probabilities via a softmax operation.

(2) Regression Head for Trajectory Generation

The regression head generates precise coordinate predictions by combining selected trajectory-mode features with social interaction context through the hierarchical decoder. This stage-wise approach ensures that only the most relevant motion patterns undergo computationally intensive social interaction modeling.

During training, the regression head processes the closest trajectory-mode feature $c_{closest}^{enc}$, determined by the minimum Euclidean distance between ground truth and trajectory-mode patterns. This closest mode feature is enhanced through social interaction modeling using Eq. (21). The final trajectory prediction is generated using a linear layer in Eq. (22).

$$c_{closest}^{dec} = \text{HierarchicalDecoder}(c_{closest}^{en}, N^{neig}, M^{neig}) \tag{21}$$

$$\hat{y} = \text{Linear}_{\text{reg}}(c_{closest}^{dec}) \tag{22}$$

During inference, the process scales to multiple modes. The top-k trajectory-mode features $\{c_i^{enc}\}_{i=1}^{k_{top}}$ selected by the classification head undergo parallel social interaction decoding, producing $k_{top}$ diverse trajectory predictions $\{\hat{y}_i\}_{i=1}^{k_{top}}$ with corresponding confidence scores.

*2.4.2 Model training strategies*

(1) Classification Loss with Soft Labeling

The classification component employs soft labeling to capture nuanced relationships between trajectory-modes and ground truth trajectories. For each training sample, soft labels $L_{soft} = \{l_k\}_{k=1}^{K}$ are computed based on trajectory similarity using Eq. (23).

$$l_k = \text{softmax}(-\delta_k), \qquad \delta_k = \|y_{gt} - m_k\|_2 \tag{23}$$

where $y_{gt}$ represents the ground-truth trajectory and $m_k$ denotes the $k$-th trajectory-mode, and $\delta_k$ is the squared Euclidean distance. This soft labeling approach enables the model to learn graduated preferences for similar trajectory-modes rather than enforcing hard binary selections.

The classification loss utilizes cross-entropy with soft labels using Eq. (24). This formulation encourages the model to assign higher probabilities to trajectory-modes that closely match the ground truth while maintaining sensitivity to alternative plausible motion patterns.

$$\mathcal{L}_{cls} = -\sum_{k=1}^{K} l_k \log(p_k), \qquad p_k = \text{softmax}(s_k^{cls}) \tag{24}$$

(2) Regression Loss for Coordinate Accuracy

The regression component is optimized using Smooth L1 loss in Eq. (25), which provides robust training dynamics for coordinate prediction. The Smooth L1 loss combines the stability of L1 loss for large prediction errors with the smooth gradients of L2 loss for small errors, promoting stable convergence and handling outliers effectively during trajectory learning.

$$\mathcal{L}_{reg} = \text{SmoothL1}(\hat{y}, y_{gt}) \tag{25}$$

The complete training objective combines both loss components with weighted strategy in Eq. (26). This joint optimization strategy enables simultaneous learning of mode selection and trajectory refinement. The classification head learns to identify relevant motion patterns from the trajectory-mode bank based on observed motion context, while the regression head focuses on precise coordinate prediction using social interaction enhancement. The weighting parameters $\lambda_{reg}$ and $\lambda_{cls}$ allow for balanced optimization between mode discrimination and trajectory accuracy, resulting in a model capable of generating diverse, socially-aware trajectory predictions with reliable confidence estimates.

$$\mathcal{L}_{total} = \lambda_{reg}\mathcal{L}_{reg} + \lambda_{cls}\mathcal{L}_{cls} \tag{26}$$

## 3 RESULTS AND DISSCUSSION

### 3.1 Experiments Settings

*3.1.1 TOD-VT dataset*

The GContextFormer model is validated using real-world vehicle trajectory data from the open-source

*Chen et al.*

TOD-VT (Video Trajectory) datase. This drone-captured dataset, developed by the ITS Research Center at Wuhan University of Technology, provides high-precision trajectory data across diverse traffic scenarios. Specifically, eight highway-ramp scenarios are used for this study as shown **TABLE 1**. All scenarios feature 3-lane mainlines with 70 km/h speed limits, while ramp configurations vary between 1-2 lanes. The dataset achieves 0.1-second temporal precision and 0.1-meter spatial precision. Raw trajectory data underwent preprocessing to generate model inputs. Frame extraction at 0.4-second intervals produces 8-step observation sequences (3.2 seconds) and 12-step prediction horizons (4.8 seconds), enabling fair comparison with existing benchmark models. The timestamp, vehicle ID, x-coordinate, and y-coordinate were extracted and converted to metric coordinates. Train-test splitting follows temporal partitioning with an 8:2 ratio to preserve neighbor interaction contexts. The processed dataset exhibits diverse traffic densities and interaction complexities across highway-ramp environments, with maximum neighbor counts varying significantly between scenarios. For detailed information about the dataset construction methodology and additional applications, readers are referred to (Wen & Lyu et al., 2024; Wen et al., 2025; Wen & Yamamoto et al., 2024).

**TABLE 1 Base Information of Highway-Ramp Scenarios**

| Set | Location | Road Type | Deceleration Lane Type | Lanes M/R | $N_{max}$ (Train / Test) |
|---|---|---|---|---|---|
| S1 | Huanle Avenue | Diverging | Parallel | 3/1 | 26 / 53 |
| S2 | Jiangcheng Avenue | Diverging | Direct | 3/1 | 48 / 50 |
| S3 | Huangpu Street | Diverging | Direct | 3/2 | 74 / 66 |
| S4 | Changfeng Avenue | Diverging | Parallel | 3/2 | 41 / 41 |
| S5 | Fazhan Avenue | Diverging | Parallel | 3/2 | 55 / 60 |
| S6 | Zhongshan Road | Merging | Direct | 3/1 | 38 / 52 |
| S7 | Jiefang Avenue | Merging | Direct | 3/2 | 28 / 32 |
| S8 | Luojiagang Road | Merging | Parallel | 3/2 | 35 / 31 |

Note: Lanes M/R denotes the number of mainline lanes (M) and ramp lanes (R), respectively. $N_{max}$ denotes, for each scenario, the maximum per-frame vehicle count observed in the training set and in the test set, respectively, computed over all time frames.

*3.1.2 Evaluation metrics*

To comprehensively evaluate trajectory prediction performance, we employ six key metrics that assess two fundamental aspects, accuracy performance and prediction robustness. All metrics are computed using the top $k_{top}$ predicted trajectories with highest confidence scores and employ the minimum error principle, selecting the trajectory with minimum error among the top $k_{top}$ most probable candidates for each test sample, reflecting practical deployment scenarios where downstream applications can utilize the most suitable prediction from multiple trajectory hypotheses.

**Minimum Average Displacement Error (minADE)** measures the average of minimum ADE values across all test samples, as defined in Eq. (27). minADE evaluates the overall trajectory accuracy by capturing temporal consistency throughout the entire prediction horizon.

$$\text{minADE} = \frac{1}{N}\sum_{n=1}^{N} \min_{i=1,\ldots,k_{top}} \frac{1}{T_{pred}} \sum_{t=1}^{T_{pred}} \|\hat{\mathbf{y}}_{n,i,t} - \mathbf{y}_{n,gt,t}\|_2 \tag{27}$$

where $\hat{\mathbf{y}}_{i,t}$ represents the predicted position at time t for the $i$-th trajectory candidate of the $n$-th test sample, $\mathbf{y}_{gt,t}$ is the corresponding ground truth position, and $N$ is the total number of test samples.

**Minimum Final Displacement Error (minFDE)** measures the average of minimum FDE values across all test samples, as shown in Eq. (28). minFDE focuses specifically on destination accuracy, which is critical for applications requiring precise endpoint prediction.

$$\text{minFDE} = \frac{1}{N}\sum_{n=1}^{N}\min_{i=1,\ldots,k_{top}}\left\|\hat{\boldsymbol{y}}_{n,i,T_{pred}} - \boldsymbol{y}_{n,gt,T_{pred}}\right\|_{2} \tag{28}$$

**Miss Rate (MR)** quantifies the prediction failure rate based on minFDE when the minimum final displacement error exceeds a specified distance threshold δ, as formulated in Eq. (29). We evaluate $MR - 2$ and $MR - 3$ with thresholds of 2 and 3 meters respectively. Miss rate assesses the model's robustness by measuring its ability to provide at least one acceptable prediction within the specified tolerance across diverse scenarios.

$$\text{MR-}\delta = \frac{1}{N}\sum_{n=1}^{N}\min_{i=1,\ldots,k_{top}}\mathbb{I}\left[\min_{i=1,\ldots,k_{top}}\left\|\hat{\boldsymbol{y}}_{n,i,T_{pred}} - \boldsymbol{y}_{n,gt,T_{pred}}\right\|_{2} > \delta\right] \tag{29}$$

where $\mathbb{I}[\,\cdot\,]$ is the indicator function that equals 1 when the condition is satisfied and 0 otherwise.

**Conditional Value at Risk (CVaR)** measures the expected minFDE of the worst-performing 20% of test samples based on their minFDE values, as defined in Eq. (30).

$$\text{CVaR} = \mathbb{E}[\text{minFDE}|\text{minFDE}\geq VaR_{80\%}(\text{minFDE})] \tag{30}$$

where $VaR_{80\%}(\text{minFDE})$ represents the 80th percentile of minFDE values across all test samples. CVaR evaluates prediction robustness by focusing on challenging scenarios where the model exhibits the poorest performance, providing insights into worst-case behavior.

### 3.1.3 Implementation details

The GContextFormer model is implemented in PyTorch and optimized using the Adam optimizer with a fixed learning rate of 1×10⁻³. The model employs a 128-dimensional hidden embedding space and sets the number of motion modes to $K = 100$. This configuration is consistent with transformer-based map-free trajectory prediction settings such as TUTR, and corresponds to the number of anchors in that setup, which keeps the training and evaluation conditions aligned for direct comparison and enables a clearer assessment of the designed modules for trajectory prediction. Training utilizes a batch size of 256 and applies a 30-meter radius for neighbor selection in social interaction masking. The architecture includes a 2-layer 4-head Motion-Aware Encoder (MAE) and a 1-layer 4-head Hierarchical Interaction Decoder (HID) with a feed-forward expansion factor of 2. Data augmentation applies mild random scaling (scale = $1 \pm 0.05$) during training. 20 trajectory samples are generated for best-of-N evaluation. This follows the prevailing configuration across map-free baselines and maintains consistent evaluation conditions so that existing methods are not disadvantaged by a different sampling protocol.

## 3.2 Model Performance Comparison
### 3.2.1 Models for comparison

To ensure fair architectural comparison, our evaluation focuses on map-free multimodal trajectory prediction methods under identical input conditions. Direct comparison with map-based approaches would conflate architectural improvements with additional input modalities. The selected benchmarks represent diverse paradigms in map-free trajectory prediction, from graph-based social modeling to transformer architectures. The following models are included in the comparison.

**STGAT (ICCV 2019)** (Huang et al., 2019): A spatial–temporal graph attention framework that models inter-agent interactions with per-step graph attention and captures temporal dependencies with recurrent encoders. Multimodality is obtained by sampling multiple trajectories at inference for best-of-N evaluation.

**Social-Implicit (ECCV 2022)** (Mohamed et al., 2022): An IMLE-based (implicit maximum likelihood estimation) approach that learns an implicit distribution over future trajectories using lightweight convolutional backbones and social zoning, without explicit graph construction. Diverse samples are generated through the IMLE sampling procedure and evaluated under best-of-N.

**Social-STGCNN (CVPR 2020)** (Mohamed et al., 2020): A spatio–temporal graph convolutional model that builds distance-aware adjacency to encode social influence and predicts future positions with a

*Chen et al.*

single-pass convolutional pipeline. Multimodal forecasts are produced by sampling from learned Gaussian outputs for best-of-N evaluation.

**Multiclass-SGCN (ICIP 2022)** (Li et al., 2022): A sparse graph convolutional architecture that incorporates agent-class labels and velocity features, employing an adaptive interaction mask to construct sparse social graphs. Multimodality is handled via probabilistic decoding with standard best-of-N evaluation.

**TUTR (CVPR 2023)** (Shi et al., 2023): A unified transformer that integrates mode discovery and trajectory decoding within an encoder–decoder design. A bank of general motion modes (anchors) is maintained, and the model jointly predicts trajectories and mode probabilities with anchor selection at inference, serves as the most relevant architectural baseline for our approach.

In addition to the baselines, we include two encoder–decoder variants. **G-HID** combines a traditional transformer encoder with the proposed hierarchical interaction decoder (HID). **G-MAE** combines the proposed motion-aware encoder (MAE) with a traditional transformer decoder. **GCF** denotes the full GContextFormer integrating both modules and is the primary model name used in tables. This benchmark selection isolates the contributions of our proposed components from additional data modalities, providing clear insights into context-aware motion encoding and hierarchical social interaction modeling effectiveness.

*3.2.2 Prediction performance comparison*

The comparative analysis covers highway-ramp scenarios from the TOD-VT dataset and reports minADE and minFDE under the unified protocol described in Section 3.1.3. All metrics are computed on the test split following the same training and inference settings. Table 2 summarizes the results across eight scenarios and shows that GContextFormer achieves the lowest mean errors (0.63 / 1.25 meters) and the highest Best Performance Ratio (63% / 88%), demonstrating superior capability in maintaining both trajectory consistency and final position precision.

**TABLE 2 Comparative results across nine scenarios: minADE / minFDE (meters)**

| Scenarios | Baseline Models | | | | | Our Models | | |
|---|---|---|---|---|---|---|---|---|
| | STGAT | Social-Implicit | Social-STGCNN | Multiclass-SGCN | TUTR | G-HID | G-MAE | GCF |
| S1 | 0.97 / 2.08 | 1.10 / 2.26 | 1.04 / 1.86 | 0.88 / 1.82 | 0.74 / 1.70 | 0.82 / 1.85 | 0.71 / 1.55 | **0.66 / 1.46** |
| S2 | 0.56 / 1.27 | 0.84 / 1.57 | 0.71 / 1.29 | 0.66 / 1.37 | 0.67 / 1.53 | 0.74 / 1.66 | 0.56 / 1.07 | **0.54 / 1.01** |
| S3 | **0.56** / 1.28 | 0.86 / 1.66 | 0.94 / 1.39 | 0.75 / 1.57 | 0.67 / 1.30 | 0.62 / 1.16 | 0.63 / **1.14** | 0.66 / **1.14** |
| S4 | 0.69 / 1.58 | 0.96 / 2.01 | 1.00 / 1.81 | 0.82 / 1.74 | 0.78 / 1.82 | 0.83 / 1.87 | 0.77 / 1.47 | **0.67 / 1.35** |
| S5 | 0.65 / 1.43 | 1.01 / 2.20 | 0.90 / 1.71 | 0.85 / 1.79 | 0.65 / 1.40 | 0.67 / 1.40 | 0.62 / **1.04** | **0.57** / 1.14 |
| S6 | 0.56 / 1.18 | 0.90 / 1.64 | 0.83 / 1.35 | 0.62 / 1.23 | 0.56 / 1.03 | 0.68 / 1.31 | 0.54 / 1.09 | **0.53 / 0.93** |
| S7 | **0.59** / 1.30 | 1.03 / 2.00 | 1.02 / 1.62 | 0.69 / 1.35 | 0.64 / 1.27 | 0.63 / 1.32 | 0.65 / 1.32 | 0.61 / **1.20** |
| S8 | **0.65** / 1.37 | 0.98 / 1.82 | 0.75 / **1.33** | 0.78 / 1.57 | 0.83 / 1.97 | 0.73 / 1.56 | 0.75 / 1.46 | 0.79 / 1.77 |
| Mean | 0.65 / 1.44 | 0.96 / 1.90 | 0.90 / 1.54 | 0.77 / 1.58 | 0.69 / 1.50 | 0.71 / 1.52 | 0.65 / 1.27 | **0.63 / 1.25** |
| BPR | - | - | - | - | - | 0% / 25% | 63% / 63% | **63% / 88%** |

Notes: All entries report the pair minADE/minFDE in meters, computed on the test set; Mean denotes the average performance across the nine scenarios/locations for each model; BPR (Best Performance Ratio) indicates, for each of our models, the ratio of scenarios in which it achieves the best performance against the five baselines, reported as "minADE win ratio / minFDE win ratio".

G-MAE attains the second-lowest mean errors (0.65 / 1.27 meters) and demonstrates strong scenario-level performance with 63% wins in both minADE and minFDE, indicating substantial gains when the motion-aware encoder replaces standard transformer encoding. In contrast, G-HID exhibits higher mean errors (0.71 / 1.52 meters) and achieves no minADE wins with only 25% minFDE wins, suggesting that hierarchical interaction decoding alone cannot compensate for the limitations of standard transformer encoding in capturing motion-specific patterns. The performance gap between G-MAE and G-HID reveals the complementary nature of the two proposed components, where motion-aware encoding provides essential trajectory understanding that enhances the effectiveness of hierarchical interaction decoding in the full GContextFormer model. Notably, the consistent performance of G-MAE across diverse scenarios underscores the fundamental importance of motion-aware trajectory-mode association, while the significant improvement in final displacement error achieved by the complete model (BPR increasing from 63% to 88% for minFDE) validates the synergistic benefits of integrating specialized motion encoding with hierarchical interaction modeling.

The analysis of baseline methods reveals distinct architectural strengths and limitations that illuminate the design rationale of GContextFormer. STGAT achieves strong performance in certain scenarios such as S3, S6, and S7, where the nature of vehicle interactions and pathway curvature align with its spatial attention mechanism. Nevertheless, its reliance on LSTM-based temporal encoding leads to notable drops in accuracy in scenarios like S1, where the prediction horizon traverses complex ramps with sharp changes in direction. This may be attributed to the difficulty of LSTM architectures in modeling long-range dependencies and adapting to abrupt motion transitions, which often result in vanishing gradients or ineffective integration of sequential spatial cues. Social-Implicit maintains consistently high prediction errors across all evaluation scenarios, averaging 0.96 and 1.90 meters for minADE and minFDE respectively. While the model leverages IMLE training to bypass adversarial loss instability typical of GAN-based methods, its minimalistic convolutional backbone and simplified pooling-based interaction mechanism substantially limit its ability to capture the nuanced multi-vehicle dynamics present in highway environments.

Graph convolutional approaches display scenario-dependent behavior. Social-STGCNN, for example, excels in S8 where the roadway geometry consists predominantly of straight multi-lane segments with minimal curvature variations, achieving the lowest error among all models in that scenario. This suggests that fixed graph topologies can be highly effective in environments where the spatial relationships between vehicles are relatively static and regular. However, such strategies lack the flexibility to adapt to more heterogeneous and dynamic scene geometries, limiting their generalizability across the wider set of highway situations. TUTR, the transformer-based baseline, achieves average minADE and minFDE scores of 0.69 and 1.50 meters. While transformer architectures provide robust global sequence modeling, TUTR employs generic transformer encoding without global context conditioning across trajectory-mode associations. This makes it less sensitive to the subtle temporal cues and vehicle-specific behaviors that are prominent in ramp and weaving zones. The substantial improvements exhibited by G-MAE over TUTR, from 0.69/1.50 to 0.65/1.27 meters, directly reflect the impact of replacing standard transformer encoding with motion-aware feature extraction. By focusing on the association between time-evolving trajectory segments and latent mode patterns specific to vehicle maneuvers, G-MAE provides richer contextual information to the decoder. This design leads to consistently superior accuracy in both trajectory and endpoint prediction, supporting the advantage of integrating specialized motion-aware encoding over generic sequence modeling in complex highway environments.

### 3.3 Modules Comparison with Transformer-based Model
*3.3.1 Performance comparison*

To isolate the contributions of our proposed modules, we conduct ablation studies using TUTR as the baseline transformer architecture (referred to as "Base" in following). TUTR serves as an ideal reference point because it employs the comparable encoder-decoder paradigm and motion mode framework as GContextFormer, enabling direct assessment of how motion-aware encoding and hierarchical interaction decoding improve upon standard transformer components. **TABLE 3** presents comprehensive quantitative results across five metrics and eight scenarios, while **Figure 4** visualizes the percentage improvements relative to the baseline, providing both absolute and relative performance perspectives for thorough analysis.

**TABLE 3 Comparative results with Transformer-based model (TUTR)**

| Scenarios | Configurations | minADE↓ | minFDE↓ | CVaR↓ | MR-2↓ | MR-3↓ |
|---|---|---|---|---|---|---|
| S1 | Base | 0.742 | 1.697 | 4.418 | 0.284 | 0.153 |
|  | G-HID | 0.817 | 1.854 | 4.562 | 0.324 | 0.168 |
|  | G-MAE | 0.709 | 1.554 | **3.968** | 0.256 | 0.129 |
|  | GCF | **0.664** | **1.461** | 4.035 | **0.221** | **0.124** |
| S2 | Base | 0.670 | 1.534 | 4.005 | 0.246 | 0.128 |
|  | G-HID | 0.743 | 1.665 | 4.339 | 0.286 | 0.156 |
|  | G-MAE | 0.561 | 1.068 | 2.989 | 0.135 | 0.066 |
|  | GCF | **0.543** | **1.015** | **2.791** | **0.115** | **0.056** |
| S3 | Base | 0.674 | 1.296 | 3.612 | 0.204 | 0.104 |
|  | G-HID | **0.616** | 1.161 | 3.217 | 0.171 | 0.082 |
|  | G-MAE | 0.629 | **1.145** | 3.086 | 0.156 | 0.073 |
|  | GCF | 0.661 | 1.145 | **3.040** | **0.155** | **0.070** |
| S4 | Base | 0.783 | 1.824 | 4.682 | 0.317 | 0.177 |
|  | G-HID | 0.825 | 1.872 | 4.821 | 0.315 | 0.179 |
|  | G-MAE | 0.769 | 1.468 | 3.763 | 0.223 | 0.112 |
|  | GCF | **0.675** | **1.354** | **3.565** | **0.195** | **0.096** |
| S5 | Base | 0.650 | 1.403 | 3.695 | 0.211 | 0.100 |
|  | G-HID | 0.669 | 1.404 | 3.811 | 0.213 | 0.110 |
|  | G-MAE | 0.620 | **1.037** | **2.900** | **0.123** | **0.057** |
|  | GCF | **0.573** | 1.138 | 3.195 | 0.154 | 0.073 |
| S6 | Base | 0.555 | 1.034 | 3.001 | 0.137 | 0.067 |
|  | G-HID | 0.679 | 1.311 | 3.502 | 0.193 | 0.098 |
|  | G-MAE | 0.540 | 1.086 | 3.162 | 0.157 | 0.079 |
|  | GCF | **0.535** | **0.926** | **2.604** | **0.104** | **0.051** |
| S7 | Base | 0.636 | 1.271 | 3.459 | 0.175 | 0.092 |
|  | G-HID | 0.635 | 1.318 | 3.733 | 0.201 | 0.107 |
|  | G-MAE | 0.646 | 1.325 | 3.543 | 0.188 | 0.098 |
|  | GCF | **0.611** | **1.204** | **3.421** | **0.174** | **0.090** |
| S8 | Base | 0.829 | 1.968 | 4.731 | 0.361 | 0.197 |
|  | G-HID | **0.728** | 1.562 | 3.933 | 0.240 | 0.124 |
|  | G-MAE | 0.750 | **1.464** | **3.760** | **0.233** | **0.117** |
|  | GCF | 0.789 | 1.768 | 4.379 | 0.302 | 0.159 |
| Mean↓ | Base | 0.693 | 1.503 | 3.950 | 0.242 | 0.127 |
|  | G-HID | 0.714 | 1.518 | 3.990 | 0.243 | 0.128 |
|  | G-MAE | 0.653 | 1.268 | 3.396 | 0.184 | 0.091 |
|  | GCF | **0.631** | **1.251** | **3.379** | **0.178** | **0.090** |
| Std.↓ | Base | 0.092 | 0.327 | 0.637 | 0.079 | 0.047 |
|  | G-HID | 0.067 | 0.226 | 0.493 | 0.049 | 0.032 |
|  | G-MAE | 0.084 | **0.202** | **0.415** | **0.046** | **0.026** |
|  | GCF | **0.084** | 0.257 | 0.545 | 0.061 | 0.034 |

Compared with the Base configuration, GContextFormer (GCF) achieves the best mean accuracy with minADE and minFDE of 0.631 and 1.251 m, improving over 0.693 and 1.503 m of Base. Consistently, miss rates and tail risk also drop substantially with CVaR from 3.950 to 3.379, MR-2 decreasing from 0.242 to 0.178, and MR-3 from 0.127 to 0.090. These trends align with the scenario-wise improvement rates in **Figure 4**, where GCF shows mean gains of 8.86% (minADE), 16.77% (minFDE), 14.47% (CVaR), 26.60% (MR-2), and 29.34% (MR-3) against Base. In terms of stability, G-MAE markedly reduces cross-scenario variance with standard deviation of minFDE decreasing from 0.327 to 0.202 (38.17%), highlighting the stabilizing effect of global context conditioning on trajectory-mode associations. GCF maintains lower variance than Base (0.257, 21.44%) while achieving the best means, indicating improved robustness across diverse highway-ramp geometries and weaving flows. Best-Performance Count analysis further corroborates these findings at the dataset level with GCF attains the highest BPC across scenarios with 6 wins in minADE, 5 wins in minFDE, 5 wins in CVaR, 6 wins in MR-2, and 6 wins in MR-3, confirming consistent superiority across diverse ramp geometries.

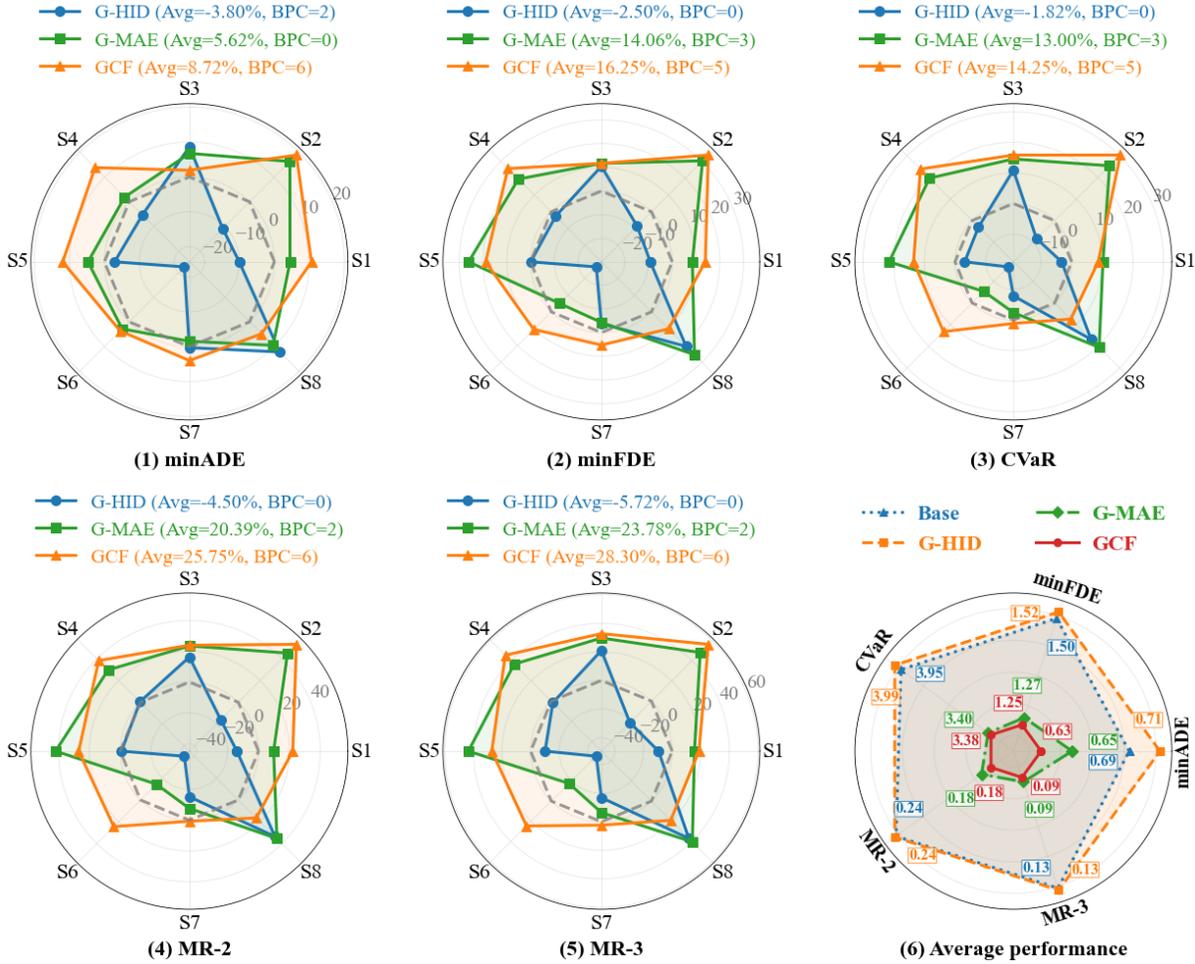

**Figure 4 Performance improvement comparison with TUTR model**

The individual module analysis reveals distinct contributions and limitations that illuminate the complementary design philosophy. Replacing the generic transformer encoder with the Motion-Aware Encoder (G-MAE) is the primary driver of gains. On average, G-MAE reduces minADE and minFDE from 0.693 and 1.503 to 0.653 and 1.268 (5.74% and 15.64%) and lowers CVaR, MR-2, and MR-3 by 14.03%, 23.96%, and 28.17%. The encoder demonstrates particularly strong performance in scenarios involving lane-changing maneuvers and continuous curvature variations, with S2 shows dramatic improvements (minFDE improvement of 30.39%, MR-2 improvement of 45.14%, MR-3 improvement of 48.20%), while S4 and S5 exhibit substantial endpoint accuracy gains of 19.52% and 26.07% in minFDE respectively. S8 also benefits significantly with 25.59% minFDE improvement and over 35% reduction in both miss rates. These improvements stem from addressing the fundamental limitation that generic transformer encoding treats trajectories as general sequence signals without trajectory-mode association conditioning. G-MAE's global context aggregation aggregates shared motion tendencies across candidate modes into scene-level

*Chen et al.*intention priors G, which suppresses spurious high magnitude activations through bounded scaled additive attention and dimension normalization thereby stabilizing the global aggregation across scenarios. The context-aware transformation mechanism then adds G to per mode queries and keys, aligning all modes in a common intention reference frame while preserving their differences through the original per mode projections. This homogeneous alignment with preserved diversity reduces cross mode homogenization and mismatches between historical observations and future mode candidates.

In contrast, introducing the Hierarchical Interaction Decoder alone (G-HID) yields scenario-dependent outcomes with mixed results. While it achieves notable improvements in S3 and S8 (minFDE improvements of 10.44% and 20.62% respectively, with S8 showing substantial miss rate reductions of 33.60% for MR-2 and 37.03% for MR-3), it exhibits degraded performance in several scenarios, particularly S1, S2, and S6, resulting in slightly worse average metrics than Base (minADE degrades by 3.09%, minFDE by 1.00%). The negative performance in S6 is particularly pronounced with minFDE degrading by 26.80% and both miss rates increasing substantially. This occurs because hierarchical interaction decoding attempts to distinguish between proximate influential neighbors and distant background vehicles, but without intention-aligned multi-modal representations as input, the hierarchical attention allocation lacks sufficient discriminative signals to function effectively. The full GContextFormer model demonstrates clear synergistic effects, where MAE provides the foundational motion-aware encoding that enables HID to deliver additional refinements in endpoint prediction and miss rate reduction. This complementary relationship is particularly evident in scenarios like S1, S4, and S6, where GContextFormer achieves further improvements over G-MAE alone. In S1, minFDE improves from 1.554 to 1.461 (an additional 5.45% beyond G-MAE's 8.46% gain), in S4 from 1.468 to 1.354 (an additional 6.25% beyond G-MAE's 19.52% gain), and most notably in S6 from 1.086 to 0.926, where the full model achieves 10.47% improvement compared to G-MAE's 4.99% degradation. Overall, the evidence shows a clear division of labor, where MAE supplies intention-aligned multi-modal encodings that cut errors and reduce variance, while HID contributes complementary interaction reasoning once fed with those motion-aware representations, together yielding the accuracy, reliability, and scenario coverage.

*3.3.2 Cases comparison across scenarios*

To illustrate the performance differences across diverse geometric configurations, **Figure 5** presents representative multimodal trajectory prediction cases from eight highway-ramp scenarios. The visualization combines aerial scene overviews with local-context magnifications and trajectory comparison plots for comprehensive analysis. The eight scenarios initially divide into two major categories based on traffic flow patterns: diverging areas (S1-S4) where vehicles separate from mainline to ramp, and merging areas (S5-S8) where vehicles integrate from ramp to mainline. These are further categorized into four specific types based on spatial characteristics: High-curvature ramp scenarios (S1, S5), Mainline-ramp transition zones (S2, S6, S7), Mainline lane change for ramp avoidance (S3), and Widening segment interactions (S4, S8). This systematic categorization enables focused analysis of how motion-aware encoding and hierarchical interaction decoding address different geometric constraints and intention disambiguation challenges inherent to highway-ramp environments. The comparative visualization reveals distinct performance patterns across model configurations, with emphasis on endpoint accuracy, multimodal candidate distribution, and alignment with actual driver trajectories.

In high-curvature ramp scenarios, S1 and S5 capturing ramp entry maneuvers where vehicles navigate sharp directional changes within confined ramp geometries. In S1, Base predictions exhibit systematic under-steering with most candidates deviating beyond the left ramp boundary, while only sparse trajectories approach the ground truth. G-MAE corrects this under-steering bias but introduces over-steering tendencies with some predictions exceeding the right boundary. G-HID achieves better convergence around the ground truth but shows limited longitudinal diversity compared to GCF's more comprehensive coverage. In S5, G-MAE increases modal diversity over Base yet again pushes some candidates beyond the right boundary, whereas G-HID turns overly conservative, tightly hugging the ground-truth arc with low spread. The complementarity of MAE (diversity with curvature awareness) and HID (precision with stability) enables GCF to balance both, keeping endpoints and local geometry well aligned with the lane centerline.

*Chen et al.*

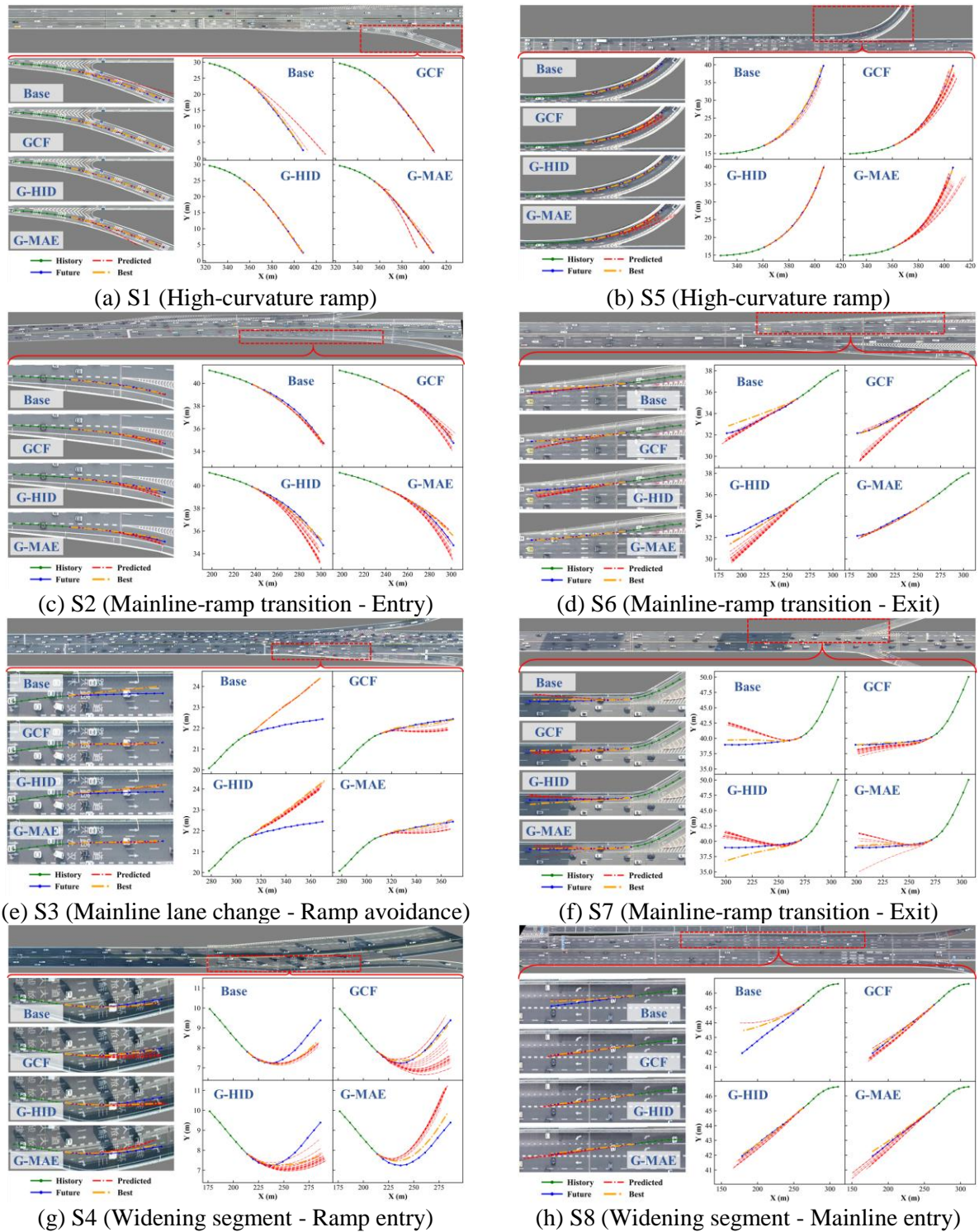

**Figure 5 Multimodal trajectory prediction on eight highway ramps**

In mainline-ramp transition zones, S2 represents entry while S6 and S7 represent exit with different geometric challenges. Although the Base endpoint lies near the ground truth, its entire family of trajectories follows an ill-shaped trend that mismatches the lane geometry. Both G-MAE and G-HID demonstrate improved dispersion, but their aggressive right-side modes cross the ramp boundary and their best candidates still mismatch the ground-truth shape. GCF achieves balanced bilateral distribution without boundary violations and superior best-prediction alignment with ground truth curvature. In S6, Base's best path under-rotates and all other hypotheses remain nearly straight, revealing weak steering anticipation in straight-going contexts. G-MAE nearly collapses the set onto the ground-truth vicinity; G-HID spreads well but skews left, including its best mode. GCF expresses two plausible modes—a straight-ahead continuation (reasonable for some drivers executing consecutive lane changes) and a right-biased path that matches the ground truth—offering both realism and accuracy. In S7, also northeast-to-west ramp exit but with direct angular connection requiring more pronounced turning, Base demonstrates over-steering with most predictions exceeding the mainline's right boundary, while G-HID produces bimodal distribution with ground truth gap and left-biased best prediction. G-MAE shows bilateral distribution with less extreme right-side trajectories and better ground truth alignment, while GCF achieves optimal performance with trajectories distributed across ground truth center and left regions (appropriate given right boundary constraints) and excellent best-prediction alignment.

S3 differs from all other cases because the target is a mainline vehicle proactively changing left to remain on the mainline when approaching a widening segment. Here, both Base and G-HID exhibit over-steering to the left and keep nearly colinear with the historical heading (G-HID is slightly more dispersed than Base), failing to reproduce the subtle, curved lane-change geometry. In contrast, G-MAE and GCF generate trajectory shapes that resemble the ground-truth's non-linear geometry, with GCF achieving well endpoint accuracy while G-MAE shows a tendency toward overly aggressive continued lane change. From a cross-scenario perspective, this distinctive case confirms that GCF adapts not only to curved ramp geometries but also effectively captures mainline lane-changing maneuvers, demonstrating broader applicability beyond ramp-specific contexts. In the widening segment category, S4 (two-lane widening before ramp entry) features continuous rightward lane changes followed by steering "release" to re-align with the lane. Base under-predicts this composite motion with line-shape errors in both longitudinal and lateral dimensions; G-MAE matches the overall shape but overdoes the late steering release; G-HID is more dispersed yet, like Base, under-releases; GCF covers both delayed and early release hypotheses, yielding a best mode that matches the ground truth. S8 (two-lane continuous leftward merge from widening to mainline) stresses sequential lane changes executed in a single maneuver. Base misclassifies it as a single lane change; G-HID's best is acceptable but the overall set is left-biased; G-MAE produces excessive longitudinal spread with premature terminal placement on the right; GCF provides a balanced longitudinal distribution, leading to a best trajectory that aligns with the intended mainline settling.

*3.3.3 Spatial distribution analysis*

While aggregate metrics provide overall performance comparisons, they may obscure spatial patterns where different models excel or struggle within specific roadway regions. To investigate these localized performance variations, we conduct spatially-resolved distribution analyses using heatmap visualizations that reveal how prediction accuracy varies across different geometric positions within each scenario. For minADE distribution analysis, we visualize performance improvements spatially anchored at observed trajectory endpoints, where heat intensity represents the improvement magnitude in minADE values relative to the Base for trajectory samples originating from those locations, revealing where trajectory evolution accuracy and early intent inference improve. Similarly, for minFDE distribution analysis, we anchor visualizations at ground-truth future trajectory endpoints, with heat intensity indicating improvement magnitude in minFDE values relative to the Base for samples terminating at those positions, isolating where terminal landing and mode selection improve. The pointwise differences are rasterized onto aerial map backgrounds via Gaussian kernel density estimation to obtain continuous heatmaps. Thus, heat values greater than zero indicate an advantage over the Base, with the magnitude representing the absolute improvement. Within each scenario panel, all three heatmaps share a common color scale to enable direct comparison of improvement magnitude and spatial coverage, highlight performance hotspots and cold zones across the road infrastructure.

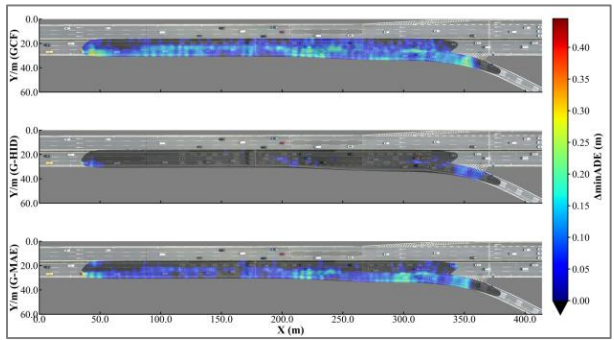
(a) S1 (Diverging area)

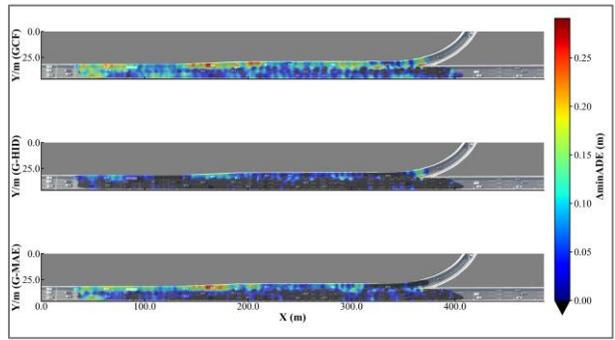
(b) S5 (Diverging area)

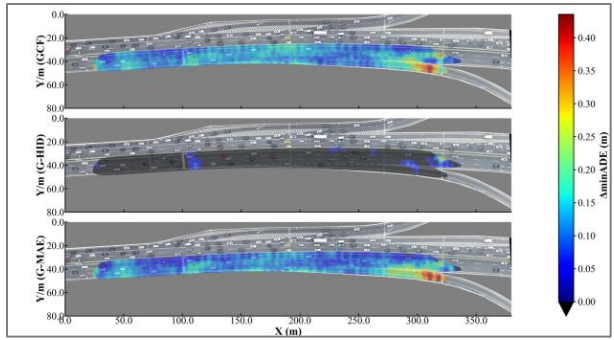
(c) S2 (Diverging area)

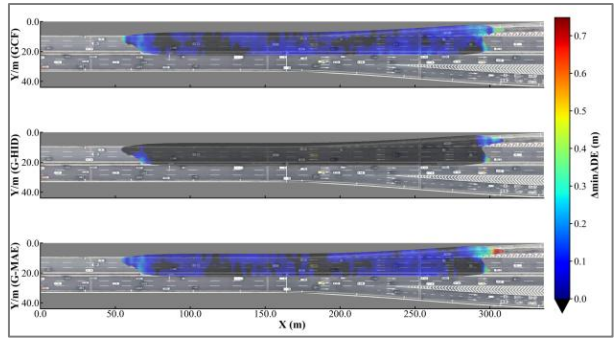
(d) S6 (Merging area)

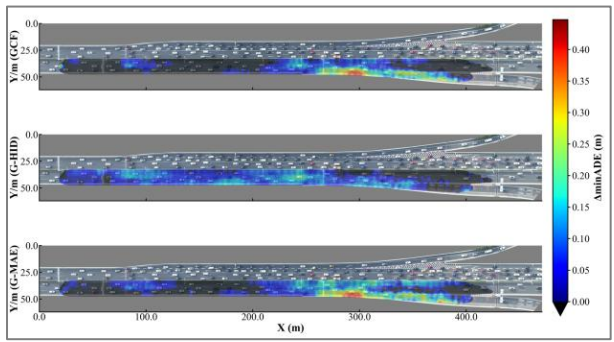
(e) S3 (Diverging area)

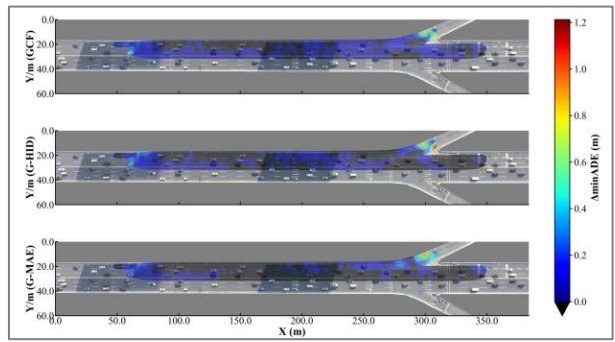
(f) S7 (Merging area)

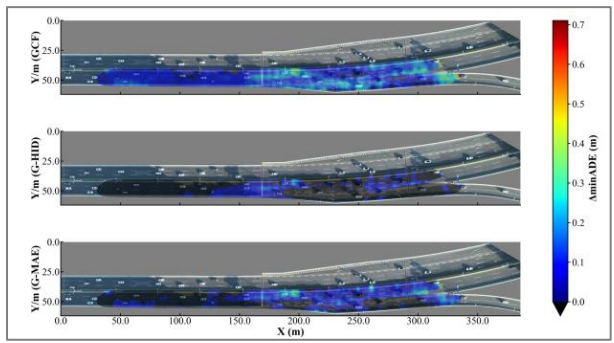
(g) S4 (Diverging area)

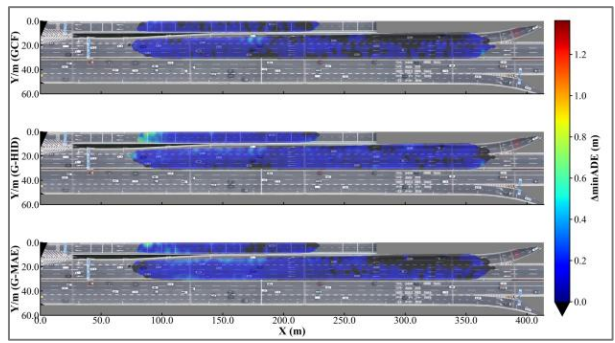
(h) S8 (Merging area)

**Figure 6 minADE distribution on eight highway ramps**

(1) Spatial distribution for minADE

**Figure 6** presents minADE improvement distributions across eight highway-ramp scenarios, revealing distinct spatial patterns in model performance. The heatmaps consistently demonstrate that GCF achieves the most widespread and intensive improvements, with warm-colored regions (indicating substantial error reductions) covering broader areas than G-HID or G-MAE alone. In high-curvature ramps (S1, S5), GCF displays the broadest and most coherent positive bands from approach to apex, indicating robust early curvature alignment and stabilized turn-in behavior. Particularly significant improvements occur at critical widening transition initiation zones (S1) and completion zones (S5), as well as initial forced diverging areas (S1) and initial merging zones (S5), where Base models typically struggle with under-rotation or exit flaring. In contrast, G-MAE shows strong but more localized positive patches near curvature onset, consistent with effective intent alignment but occasional over-aggressive modes that reduce overall spatial coverage, while G-HID displays more localized improvements with irregular spatial distribution. Transition zone scenarios (S2, S6, S7) reveal GCF's superior performance in resolving intent disambiguation challenges, with intense improvement regions concentrated at decision points where vehicles commit to ramp entry or exit. In S2, GCF forms a continuous high-gain strip that spans the feeder into the ramp nose and the immediate post-entry alignment. In S6 and S7, GCF covers both the mainline-continuation and ramp-bound corridors, reflecting robust disambiguation of competing intents.

In ramp-avoidance scenarios (S3), GCF and G-MAE present clear positive strips along the subtle non-linear lane-keeping path, while G-HID's gains appear narrower spatially. This narrower improvement pattern for G-HID reflects its tendency toward near-collinearity with observed history, which enables it to achieve positive improvements across most straight-driving regions. This reduces error in straight segments but provides less benefit in the small yet safety-critical lateral adjustments around the widening section and the ramp throat. Indeed, although G-HID attains a slightly lower aggregate minADE than GCF in S3, the heat-map reveals that its gains are concentrated on straight, low-interaction stretches, whereas GCF and G-MAE deliver larger benefits precisely in the non-linear interaction areas that are more relevant to collision avoidance. In the widening scenarios S4 and S8, GCF again produces physically meaningful improvement corridors. In S4, warm bands appear both in the gradual lateral drift zone and in the subsequent steering-recovery zone where vehicles complete the alignment after the drift, indicating that GCF corrects both early and delayed timing. For S8, GCF clusters improvements at the lane-settling area into mainline, mitigating Base's auxiliary-lane persistence. Although G-MAE and G-HID show slightly better aggregate minADE performance, the spatial analysis reveals that their significant improvement regions are located on the independent auxiliary roadway above the main highway, which does not constitute social interaction with main highway vehicles. In contrast, GCF's significant improvement regions are positioned in the latter half of the two-lane widening segment after ramp exit, where vehicles have begun or are preparing to accelerate for mainline entry. This finding represents insights that aggregate numerical results cannot capture.

(2) Spatial distribution for minFDE

**Figure 7** focuses on minFDE improvements over the Base, complementing the minADE results by emphasizing endpoint landing precision rather than trajectory evolution. In high-curvature scenarios (S1, S5), GCF produces the most concentrated and intense positive clusters at curve completion zones, indicating superior endpoint sharpening. These concentrated heat regions demonstrate GCF's ability to converge predicted trajectories toward correct terminal positions with higher precision than Base. G-MAE shows substantial endpoint improvements distributed along the curve's terminal arc but with less concentration at specific landing points, while G-HID provides localized improvements with smaller spatial coverage and lower intensity values. In transition zones, for S2 entry scenarios, GCF eliminates the scattered underperforming regions around ramp junction points that persist with G-MAE and G-HID, achieving more uniform positive improvements across potential entry points and indicating more decisive mode selection at entry commitment moments. For S6 and S7 exit scenarios, GCF concentrates improvements precisely at valid exit terminal positions while suppressing errors associated with adjacent or wrong lane predictions that affect Base performance. The heat distributions show GCF maintains positive improvements across multiple plausible exit endpoints, while G-MAE shifts improvement mass toward correct lanes but with less terminal precision, and G-HID often exhibits spatial splitting.

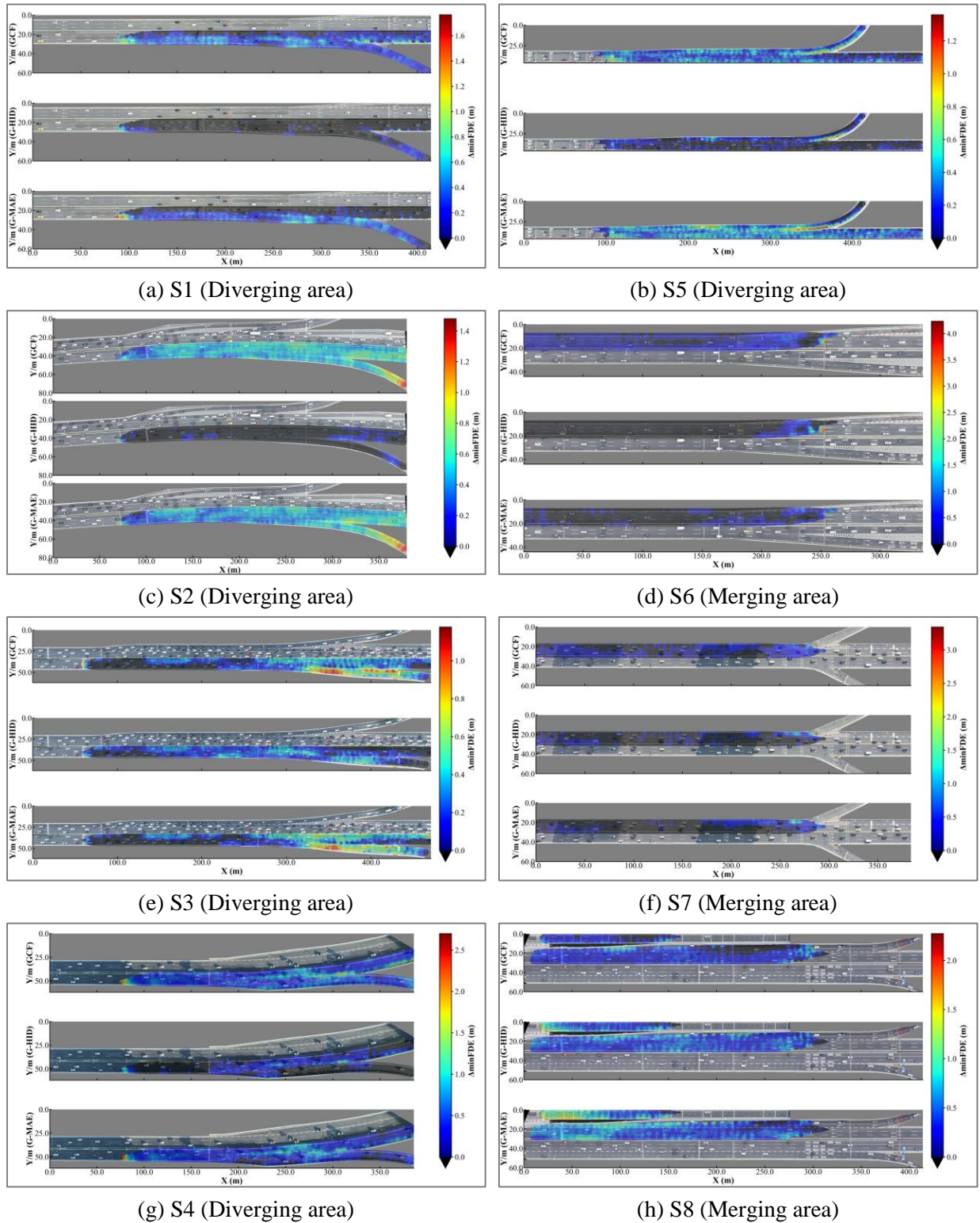

(a) S1 (Diverging area)  (b) S5 (Diverging area)

(c) S2 (Diverging area)  (d) S6 (Merging area)

(e) S3 (Diverging area)  (f) S7 (Merging area)

(g) S4 (Diverging area)  (h) S8 (Merging area)

**Figure 7 minFDE distribution on eight highway ramps**

In ramp-avoidance scenarios (S3), GCF shows concentrated improvements in the terminal regions where the gentle S-shaped lane-keeping maneuver reaches completion, with positive heat concentrated at the expected endpoint region. This indicates superior prediction of subtle maneuver completion compared to Base's tendency toward linear extrapolation. The scenario features a ramp segment that first curves upward (with center point above) then curves downward (with center point below). G-MAE achieves similar but slightly less focused improvements, showing weakness in the latter half of the ramp segment at the initial downward curve position compared to GCF, while G-HID performs most modestly with only marginal improvements over Base. In widening segments S4 and S8, GCF markedly improves final alignment in regions where vehicles complete steering corrections to stabilize in target lanes. For S4, GCF demonstrates concentrated positive improvements at appropriate terminal positions, capturing both early and late steering recovery outcomes effectively. For S8, the scenario appears distinctive with extensive straight-line segments where GCF's performance may be slightly weaker than other models in simple trajectory continuation. However, GCF's most significant improvements are positioned at the end of the auxiliary roadway from ramp exit, specifically at the roadway narrowing transition zone where auxiliary road vehicles must merge into the main roadway. While G-HID and G-MAE also perform excellently in this scenario, suggesting that simpler models like G-MAE may be more suitable for scenarios dominated by straightforward linear motion, GCF's superior performance in nonlinear motion prediction remains undeniable.

Overall, the spatial distribution analysis reveals complementary strengths. G-MAE drives substantial improvements where early intent establishment and curvature consistency matter most, providing the foundation for better trajectory evolution. G-HID adds localized refinements along ground-truth corridors but with limited spatial coverage when driving intent remains ambiguous. GCF combines and amplifies both strengths, producing extensive minADE improvements across approach and evolution phases while delivering concentrated and precise minFDE improvements at geometric decision points and terminal positions, particularly excelling in complex nonlinear motion scenarios that are critical for real-world autonomous driving applications.

## 3.4 Approach Potential

*3.4.1 Model inference explanation*

Our GContextFormer provides built-in interpretability by exposing attention signals at each reasoning stage and by aligning these signals with geometry- and interaction-aware priors. The model decomposes inference into two complementary pathways, a mode-aware encoder (MAE) that weighs global motion prototypes against the ego's observation, and a human–interaction decoder (HID) that assigns attention to surrounding agents and fuses their influence through cross-attention. To extract attention patterns, we perform forward inference on a highway ramp scenario and capture intermediate attention weights from both the Motion-Aware Encoder (MAE) and Hierarchical Interaction Decoder (HID) components. For MAE, we treat the library of general motion modes as a fixed geometric carrier and color each mode by the corresponding attention weights, which are extracted from both global context aggregation layers and their context-aware interaction layers, averaging across attention heads to obtain trajectory-mode attention scores, maintaining a common color scale across layers of the same submodule. These scores indicate how strongly each predefined motion mode aligns with the observed trajectory context. For HID, we analogously visualize the weights over neighbor context attention and dual-pathway cross-attention matrices, where neighbor context attention reveals the global importance of each surrounding vehicle, and cross-attention matrices show fine-grained dependencies between trajectory modes and social agents. The visualization pipeline renders these internal weights back into the physical coordinate frame so that each attention pattern can be read as "which motion hypothesis" and "which neighbor" drove the prediction.

(1) Motion-Aware Encoder analysis

The MAE performs a two-stage, coarse-to-fine screening of motion hypotheses. The global mode-context branch first filters the prototype set by geometric compatibility with the observed history; the context-aware transformation branch then re-weights the shortlisted modes by how well they align with the evolving latent context. The layer-wise patterns exhibit complementary functions where the first layer explores and gates, the second layer sharpens and calibrates, and the layer-aggregated maps retain multimodality while elevating the most plausible mode.

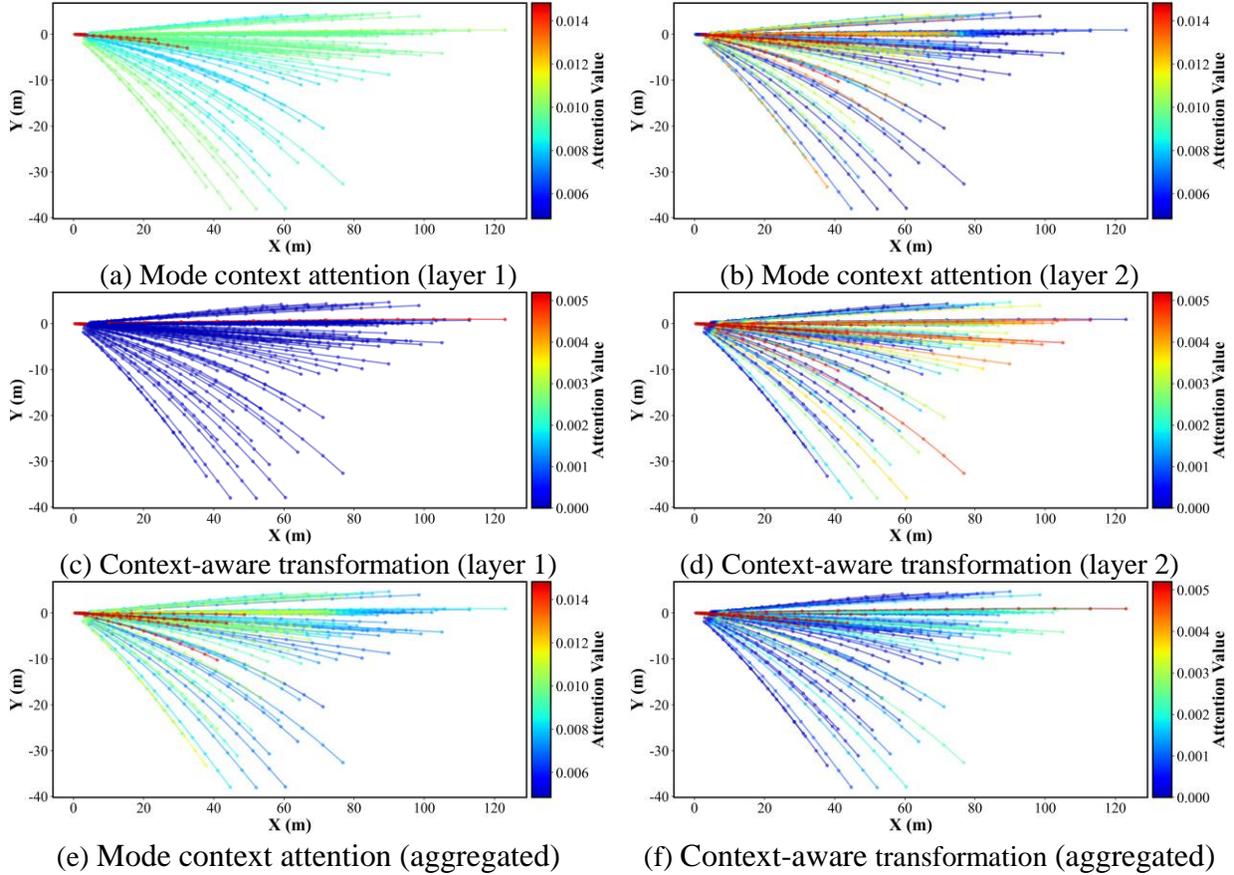

(a) Mode context attention (layer 1)      (b) Mode context attention (layer 2)

(c) Context-aware transformation (layer 1)      (d) Context-aware transformation (layer 2)

(e) Mode context attention (aggregated)      (f) Context-aware transformation (aggregated)

**Figure 8 Averaged attention distributions for MAE**.

**Figure 8** (a) and (b) present mode context attention across MAE's two encoding layers, revealing the model's evolving motion pattern prioritization. In Layer 1 (**Figure 8** (a)), attention distribution appears relatively concentrated, with most modes receiving light blue values and light green values, while only a few modes achieve red coloring indicating peak attention. This concentrated pattern reflects initial geometric filtering where the model identifies motion modes that align with the observed trajectory's basic directional and curvature characteristics. Layer 2 (**Figure 8** (b)) demonstrates markedly different attention dynamics, with colors spanning the entire spectrum from deep blue to red. The expanded color distribution indicates enhanced discriminative capacity, where the second layer applies more sophisticated contextual analysis to either suppress geometrically incompatible patterns (deep blue) or enhance contextually relevant ones (warm colors). This progression from concentrated to diverse attention reveals the model's ability to move beyond simple geometric matching toward nuanced contextual understanding of trajectory-mode associations. The aggregated mode context attention (**Figure 8** (e)) balances the concentrated exploration of Layer 1 with the discriminative refinement of Layer 2. Compared to Layer 1, the aggregated attention displays more red modes and deeper blue modes, along with some yellow-colored modes, while reducing the prevalence of the deepest blue modes seen in Layer 2. This aggregation achieves a refined attention distribution that maintains multimodal awareness while focusing on contextually appropriate patterns, demonstrating how the two layers work complementarily to establish motion pattern priorities.

**Figure 8** (c) and (d) illustrate context-aware interaction attention across MAE's transformation layers, showing how motion modes are refined through interaction with global trajectory context. Layer 1 (**Figure 8** (d)) exhibits a distinct bipolar attention pattern with modes concentrated at color spectrum extremes. This "gating" behavior promotes modes that best agree with the latent context formed by the history while strongly down-weighting incompatible ones. Layer 2 (**Figure 8** (d)) presents a more distributed attention pattern spanning the full spectrum, similar to mode context attention Layer 2 but with reduced deep blue presence and increased light yellow and light red colors in the middle range. This suggests that context-aware transformation operates through gradual contextual integration rather than binary keep/drop, fine-tunes relative preferences among already plausible modes. The aggregated context-aware transformation attention (**Figure 8** (f)) shows characteristics like the mode context aggregation but the transformation aggregation features more deep blue modes, substantial light blue and blue-green modes, and concentrated warm colors. This pattern reveals that context-aware interaction operates complementarily to mode context attention, while mode context attention establishes geometric feasibility, context-aware transformation ensures behavioral consistency with observed trajectory evolution patterns. This indicates conservative preservation of multiple feasible options while selectively elevating those most consistent with the local trajectory context.

The attention evolution across MAE components demonstrates systematic reasoning, mode context attention layers establish geometric compatibility through progressive discrimination from broad exploration to focused selection, while context-aware interaction layers ensure behavioral consistency through trajectory-context integration. This dual-pathway approach enables GContextFormer to balance geometric constraints with behavioral expectations, providing interpretable insights into how the model weighs motion pattern feasibility against contextual appropriateness in trajectory prediction tasks. This two-stage refinement is precisely what allows GContextFormer to be both accurate in curved/branching regions and robust across diverse highway geometries.

(2) Hierarchical Interaction Decoder analysis

The HID employs a dual-pathway attention mechanism that decomposes social reasoning into complementary stages: neighbor context attention establishes agent-centric, scene-level saliency priors independent of specific ego motion hypotheses, while dual cross-attention channels provide mode-conditioned, pairwise interaction assessment. This architecture creates an interpretable reasoning chain from "which neighbor matters globally" to "for which motion mode it matters specifically," enabling nuanced social interaction modeling beyond undifferentiated proximity-based weighting. The former measures global agent saliency in the traffic scene with ego included, while the latter measures pairwise constraints for each candidate ego motion.

**Figure 9** visualizes neighbor context attention through self-attention over the complete agent set, where surrounding vehicles are reordered by attention magnitude with ego serving as reference. The attention hierarchy reflects sophisticated interaction priority understanding: Neighbor 1 (light blue) represents the leading vehicle directly ahead, receiving highest neighbor attention due to immediate influence on forward trajectory planning. Neighbor 2 (green) corresponds to a left mainline vehicle, initially positioned slightly behind ego but advancing ahead due to higher speed and longitudinal overtaking dynamics relative to ego make it a salient monitoring target despite lane separation. Neighbor 3 (light yellow) represents the closely following vehicle directly behind ego, while Neighbor 4 (purple) occupies the left-rear position behind both ego and Neighbor 3. Neighbor 5 (light brown) is positioned right-rear, gradually merging into ego's lane. This ranking reflects an ego-conditioned, scene-level contextual perspective rather than direct ego–neighbor relationships. The self-attention mechanism captures individual relations to ego together with their broader effects on scene dynamics, including salient changes in local traffic structure that warrant monitoring even when the immediate interaction probability is low.

**Figure 10** (a) and (b) present the dual cross-attention pathways, where rows represent top-k trajectory modes and columns represent agents (0=ego, 1-5=neighbors). The context-enhanced cross-attention (**Figure 10** (a)) shows markedly different patterns compared to standard cross-attention (**Figure 10** (b)), revealing how global context modulates fine-grained trajectory-agent interactions. For Neighbors 2 and 5, context enhancement reduces already minimal attention to near-zero levels, demonstrating appropriate filtering of globally salient but geometrically decoupled agents. For Neighbor 1, context enhancement creates sharper trajectory-mode differentiation while redirecting some attention to ego (column 0), suggesting that different ego trajectory modes require varying levels of consideration of leading vehicle behavior. Neighbor 4 demonstrates selective attention retention primarily for trajectory-mode 0, indicating lane-specific interaction relevance. Conversely, attention redistribution amplifies interactions with geometrically relevant neighbors, Neighbor 3 (following vehicle) exhibits increased attention differentiation across trajectory modes, indicating that follower interactions vary substantially with ego's intended motion—conservative modes demand careful follower consideration while aggressive modes may not.

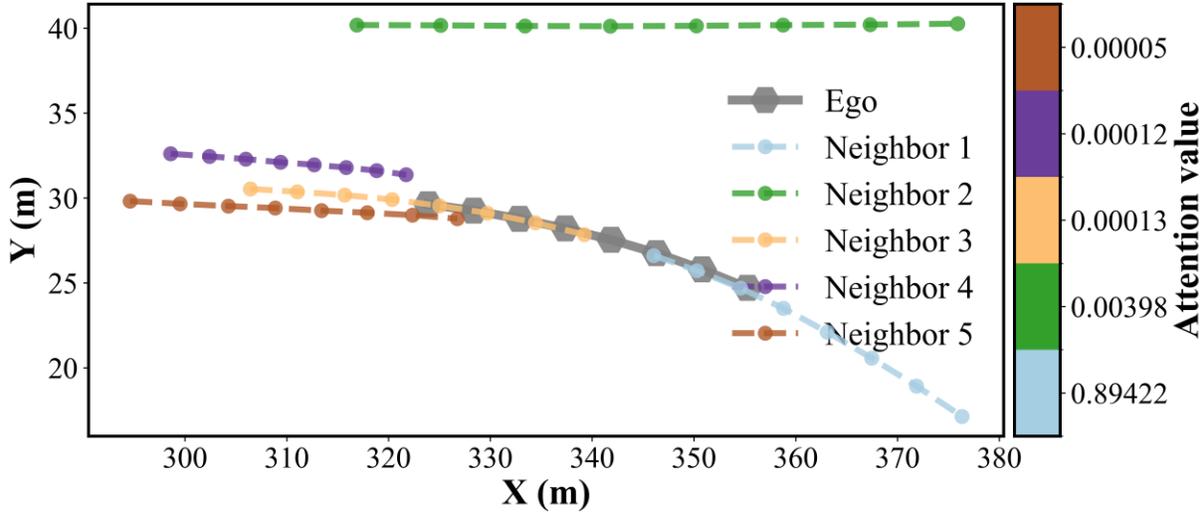

**Figure 9 Neighbor context attention distribution for HID**.

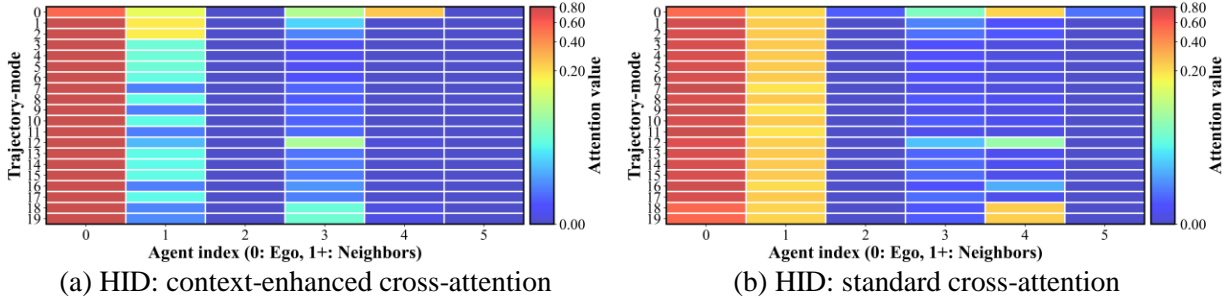

(a) HID: context-enhanced cross-attention      (b) HID: standard cross-attention

**Figure 10 The dual-pathway cross attention distributions for HID**.

With shared parameters, the dual path cross attention provides complementary coverage and balanced fusion. The standard pathway evaluates geometric relations across all pairs of agents and candidate modes in a uniform way, which preserves coverage of subtle or low priority interactions and supplies a consistent baseline. The context enhanced pathway uses the neighbor context prior from **Figure 9** to reweight these relations, reducing attention to distant or lane separated agents and sharpening focus on neighbors that truly constrain feasible motion. The gated fusion combines these signals so the model keeps broad geometric sensitivity while limiting unwanted attention. Additionally, relying on the context enhanced pathway alone can miss weak but important geometry conditioned cues and can inherit potential miscalibration in the prior, whereas shared parameters keep both pathways aligned and allow the gate to choose per scene and per mode, yielding focused and complete social reasoning for trajectory prediction.

*3.4.2 Module extensibility and potential improvements*

The proposed modular design exhibits extensibility potential within trajectory prediction and beyond. The Motion-Aware Encoder (MAE) establishes a scene-level intention prior through scaled additive aggregation over general motion modes, then refines per-mode representations under shared global context while preserving mode distinctions. Our modification to the classical Bahdanau additive attention mechanism replaces the dot-product interaction with a bounded nonlinearity over ($Q + K$) and introduces scaling normalization term $1/\sqrt{d_k}$ to prevent mode-specific bias accumulation during global aggregation. This design avoids introducing a fixed preference toward prevalent motion patterns across modes. Unlike multiplicative attention ($Q \cdot K$), which tends to amplify feature-magnitude differences, the additive formulation ($Q + K$) with bounded nonlinearity ($\tanh(\cdot)$) ensures balanced representation across heterogeneous motion modes. the tanh activation provides nonlinear selectivity while preventing score domination by high-magnitude features, and the scaling factor further stabilizes optimization. This mechanism is particularly beneficial in ramp scenarios, where prevalent straight-line patterns might otherwise suppress equally plausible turning behaviors.

The Hierarchical Interaction Decoder (HID) separates social reasoning into two parameter-shared pathways, where the standard cross-attention maintains uniform geometric coverage over agent-mode pairs, and the context-enhanced pathway reweights these couplings using neighbor-context priors that establish scene saliency. Gated fusion aligns both pathways in a common representation space and enables scene- and mode-dependent attention selection. The retention of the standard pathway serves to recover faint, geometry-conditioned influences that may be downplayed by the prior, preventing prior miscalibration or missing weak yet consequential signals. Additionally, attention maps can reveal which neighbors and which ego modes drive the prediction, facilitating interpretability and diagnostic capabilities, while offering actionable signals for risk assessment, particularly in safety-critical scenarios where understanding interaction dynamics becomes paramount for system reliability.

For broader traffic scenarios, the general motion modes, derived from normalized trajectory clustering, provide a possible transferable template that can be expanded to accommodate heterogeneous agent behaviors across diverse road environments. Highway-to-urban transfer can benefit from shared curvature patterns and lane-changing behaviors, though the integration of motion patterns from multiple agent types and road conditions typically requires expanding the mode libraries and refining transformation sensitivity to capture higher decision complexity, allowing MAE to learn adaptive representational alignment processes across such heterogeneous modes. The global context aggregation from MAE shows potential for integration into map-lite deployments where coarse topology or sparse lane cues are available, with the intention prior absorbing scene regularities while preserving per-mode diversity. In map-rich settings, the module could incorporate lane semantics as additional context features while preserving the fundamental scaled additive aggregation and refinement architecture that enables motion-mode conditioning, thereby supporting broader behavioral adaptation and prediction capabilities for mixed agent scenarios.

On the agent interaction side, the neighbor-context prior from HID can be extended from ego-centric neighborhoods to multi-scale neighborhoods encompassing crossable and non-crossable lane groups, heterogeneous agent classes, and behavioral hierarchies. In practice, priors are composed across scales and the gate mediates their relative influence within a shared representation, potentially supporting more comprehensive adaptation across highway segments and urban junctions. The dual-pathway cross-attention approach naturally accommodates heterogeneous agent types through considering agent-specific embedding layers while maintaining the core interaction reasoning. Mixed traffic scenarios involving vehicles, pedestrians, and cyclists could leverage agent-type-aware motion modes combined with the hierarchical interaction structure, where global neighbor context captures cross-agent priority while standard cross-attention preserves fine-grained behavioral coupling. Beyond spatial scale and heterogeneity, temporal reach can be increased by expanding the motion-mode set to include longer-horizon patterns while the global context mitigates mode collapse during longer-term prediction.

*Chen et al.*Overall, MAE provides intention-aligned multimodal adjustment between historical trajectories and motion modes under global context reference, while HID provides controllable trade-offs between coverage and focus in neighbor interactions. The modules constitute a plug-and-play stack through complementary design principles rather than rigid architectural coupling, which enables each to be extended or adapted independently when scenarios, priors, or map granularity change.

*3.4.3 Cross-Domain Application Prospects*

The motion-aware encoding paradigm extends to transportation tasks that require multimodal reasoning under uncertainty. At the network scale, traffic flow prediction could utilize the global context mechanism to aggregate flow patterns across multiple intersections or highway segments, with individual link predictions conditioned on network-wide traffic states. The bounded scaled additive attention mechanism ensures stable aggregation across varying node counts, while the context-aware transformation enables local flow adaptations based on global traffic conditions. Route planning applications could leverage the trajectory-mode association framework where motion modes represent different routing strategies rather than vehicle trajectories. The global context would capture network constraints and traffic conditions, while individual route modes maintain strategic preferences for time-minimization, energy-efficiency, or comfort optimization. The hierarchical interaction decoder could model multi-vehicle coordination in autonomous vehicle platoons, where neighbor context establishes formation priorities and cross-attention pathways handle vehicle-specific coordination signals.

These architectural principles also show potential beyond transportation where uncertainty, multimodality, and spatial interactions are prominent. In human motion synthesis, scene geometry and social norms can provide the global context while modes capture activity- or style-specific motion. The scaled additive attention mechanism would ensure stable aggregation across varying numbers of people or activity types, though the temporal scales and interaction patterns would require careful domain-specific adaptation. Swarm robotics applications may benefit from global neighbor context to identify formation leaders or coordination anchors, with dual-pathway interactions handling task-specific inter-robot communications. The masking mechanism would naturally accommodate dynamic swarm membership changes, though real-world deployment would need to address communication constraints and latency considerations. Multi-party collaborative systems including distributed sensing networks could potentially apply these principles where global context establishes communication priorities while cross-attention enables task-specific information exchange between sensor nodes. These cross-domain applications represent conceptual extensions rather than validated implementations. Each requires careful treatment of data characteristics, temporal scales, and interaction patterns. Such extensions would require domain-specific modifications to embedding schemes and training objectives while preserving the core architectural insights of global context conditioning for intention alignment and hierarchical interaction for coverage–focus balance.

# 4 CONCLUSIONS

Given that existing multimodal trajectory prediction approach often rely on HD maps that are vulnerable to incomplete coverage, stale updates, or even adversarial errors. In contrast, those operating without such references struggle to maintain global motion–intention consistency across multiple plausible futures, as conventional pairwise attention mechanisms over-amplify dominant straight patterns while suppressing minority transitional patterns (i.e., merging, diverging and curved ramp transitions). This study proposes GContextFormer, a plug-and-play encoder–decoder architecture designed to achieve intention-aligned multimodal prediction and balanced social reasoning without explicit map dependence. Its modular design forms a complementary stack in which each component can function independently or be extended when contextual information, prior knowledge, or map granularity changes. The architecture provides context-grounded interpretability and scalability as a general foundation for future trajectory-reasoning models.

GContextFormer unifies global intention understanding and hierarchical social interaction modeling through two synergistic components, mitigating the bias of pairwise attention toward frequent geometrically

*Chen et al.*aligned dominant motion patterns. The Motion-Aware Encoder (MAE) establishes a scene-level intention prior via bounded scaled additive aggregation, promoting coherent context sharing among trajectory-mode representations while mitigating inter-mode suppression. The Hierarchical Interaction Decoder (HID) introduces a dual-pathway attention mechanism comprising distinct yet complementary reasoning streams. The neighbor-context-enhanced pathway constructs an agent-centric saliency prior that captures scene-level relevance among neighboring agents under a shared neighbor-context prior, emphasizing salient interactions beyond mere proximity. In contrast, the standard cross-attention pathway provides uniform geometric coverage over agent-mode pairs, ensuring consistent interaction reasoning across spatial distributions. A learnable gating fusion adaptively balances global saliency focus and local interaction coverage, forming an interpretable reasoning chain from which neighbor matters globally to for which motion mode it matters specifically.

Extensive evaluations on eight highway-ramp scenarios show that GContextFormer achieves the highest overall accuracy among state-of-the-art benchmark models, delivering the lowest cross-scenario average minADE and minFDE and the highest best-performance ratios (63% / 88%). It further maintains strong robustness, reducing miss rates by up to 26.60% and 29.34% relative to transformer-based baselines. Beyond these widely-adopted aggregate metrics, performance spatial distribution analysis reveals concentrated improvements in complex nonlinear scenarios including high-curvature ramps and mainline-ramp transition zones. Moreover, interpretability emerges through motion mode distinctions and neighbor context modulation, which expose the reasoning process from intention understanding to interaction modeling. The former reveals which motion modes drives each potential trajectory among competing modes, while the latter identifies which surrounding agents matter most from global scene-level relevance to specific agent-mode impact. The architecture also demonstrates notable modularity and extensibility potential, providing a transferable framework for multimodal motion prediction under varying levels of contextual completeness. Future work will focus on hierarchical temporal reasoning for improved long-horizon forecast stability, enhanced general motion-mode learning for broader behavioral coverage, and multi-granularity context embedding, including optional integration of map or semantic cues when available. Further extending the HID toward multi-agent cooperative reasoning may facilitate stronger cross-scene generalization and unified modeling of heterogeneous interacting entities.

**ACKNOWLEDGMENTS**
The authors would like to thank Dr. Jiaqiang Wen from Wuhan University of Technology for generously sharing his insights regarding the TOD-VT dataset. This work was supported in part by the National Key R&D Program of China [grant number 2023YFE0106800], Outstanding Youth Foundation of Jiangsu Province [grant number BK20231531], Frontier Technologies R&D [Program of Jiangsu |grant number BF2024019], and the Postgraduate Research & Practice Innovation Program of Jiangsu Province [grant number KYCX23_0306, KYCX25_0510].

**AUTHOR CONTRIBUTIONS**
The authors confirm contribution to the paper as follows:

**DECLARATION OF CONFLICTING INTERESTS**
The authors declare that they have no known competing financial interests or personal relationships that could have appeared to influence the work reported in this paper.

**Data availability**
The data used in this study is publicly available on the Professor Nengchao Lyu's Research Group, ITS Research Center, Wuhan University of Technology website. Interested readers can contact Prof. Lyu and Dr. Wen directly. We do not have the authority to redistribute the data.

**References**

Bahdanau, D., Cho, K., & Bengio, Y. (2014). Neural machine translation by jointly learning to align and translate. arXiv preprint arXiv:1409.0473 http://doi.org/10.48550/arXiv.1409.0473

Carrasco Limeros, S., Majchrowska, S., Johnander, J., Petersson, C., & Fernández Llorca, D. (2023). Towards explainable motion prediction using heterogeneous graph representations. Transportation Research Part C: Emerging Technologies, 157, 104405. http://doi.org/10.1016/j.trc.2023.104405

Chen, Y., Zou, Y., Xie, Y., Zhang, Y., & Tang, J. (2025). Multimodal vehicle trajectory prediction based on intention inference with lane graph representation. Expert Systems with Applications, 262, 125708. http://doi.org/10.1016/j.eswa.2024.125708

Cheng, H., Liu, M., Chen, L., Broszio, H., Sester, M., & Yang, M. Y. (2023). GATraj: A graph- and attention-based multi-agent trajectory prediction model. ISPRS Journal of Photogrammetry and Remote Sensing, 205, 163-175. http://doi.org/10.1016/j.isprsjprs.2023.10.001

Deo, N., Wolff, E., & Beijbom, O. (2022). Multimodal trajectory prediction conditioned on lane-graph traversals. Proceedings of the 5th Conference on Robot Learning, Proceedings of Machine Learning Research.

Gu, X., Song, G., Gilitschenski, I., Pavone, M., & Ivanovic, B. (2024). Producing and leveraging online map uncertainty in trajectory prediction. 2024 IEEE/CVF Conference on Computer Vision and Pattern Recognition (CVPR), Seattle, WA, USA. http://doi.org/10.1109/CVPR52733.2024.01376

He, Y., Xie, H., & Zhang, X. (2025). DCSTNet: A dual-channel spatio-temporal information fusion network for map-free vehicle trajectory prediction. IET Intelligent Transport Systems, 19(1), e70030. http://doi.org/10.1049/itr2.70030

Huang, R., Pagnucco, M., & Song, Y. (2023). HyperTraj: Towards simple and fast scene-compliant endpoint conditioned trajectory prediction. 2023 IEEE/RSJ International Conference on Intelligent Robots and Systems (IROS), Detroit, MI, USA. http://doi.org/10.1109/IROS55552.2023.10341647

Huang, R., Xue, H., Pagnucco, M., Salim, F. D., & Song, Y. (2025). Vision-based multi-future trajectory prediction: A Survey. IEEE Transactions on Neural Networks and Learning Systems, 1-18. http://doi.org/10.1109/TNNLS.2025.3550350

Huang, Y., Bi, H., Li, Z., Mao, T., & Wang, Z. (2019). STGAT: Modeling spatial-temporal interactions for human trajectory prediction. 2019 IEEE/CVF International Conference on Computer Vision (ICCV), Seoul, Korea (South). http://doi.org/10.1109/ICCV.2019.00637

Kawasaki, A., & Seki, A. (2021). Multimodal trajectory predictions for autonomous driving without a detailed prior map. 2021 IEEE Winter Conference on Applications of Computer Vision (WACV). http://doi.org/10.1109/WACV48630.2021.00377

Lei, D., Xu, M., & Wang, S. (2025). A deep multimodal network for multi-task trajectory prediction. Information Fusion, 113, 102597. http://doi.org/10.1016/j.inffus.2024.102597

Li, G., Li, Z., Knoop, V. L., & van Lint, H. (2024). Unravelling uncertainty in trajectory prediction using a non-parametric approach. Transportation Research Part C: Emerging Technologies, 163, 104659. http://doi.org/10.1016/j.trc.2024.104659

Li, L., Wang, X., Yang, D., Ju, Y., Zhang, Z., & Lian, J. (2024). Real-time heterogeneous road-agents trajectory prediction using hierarchical convolutional networks and multi-task learning. IEEE Transactions on Intelligent Vehicles, 9(2), 4055-4069. http://doi.org/10.1109/TIV.2023.3275164

Li, R., Katsigiannis, S., & Shum, H. P. H. (2022). Multiclass-SGCN: Sparse graph-based trajectory prediction with agent class embedding. 2022 IEEE International Conference on Image Processing (ICIP), Bordeaux, France. http://doi.org/10.1109/ICIP46576.2022.9897644

Liao, B., Chen, S., Zhang, Y., Jiang, B., Zhang, Q., Liu, W., Huang, C., & Wang, X. (2025). MapTRv2: An end-to-end framework for online vectorized HD map construction. International Journal of Computer Vision, 133(3), 1352-1374. http://doi.org/10.1007/s11263-024-02235-z

Liao, H., Li, Z., Wang, C., Shen, H., Liao, D., Wang, B., Li, G., & Xu, C. (2024). MFTraj: Map-free, behavior-driven trajectory prediction for autonomous driving. Proceedings of the Thirty-Third International Joint Conference on Artificial Intelligence, Jeju, Korea. http://doi.org/10.24963/ijcai.2024/65

Liu, X., Xing, Y., & Wang, X. (2024). Map-free trajectory prediction with map distillation and hierarchical encoding. arXiv preprint arXiv:2411.10961 http://doi.org/10.48550/arXiv.2411.10961

Liu, Y., Yuan, T., Wang, Y., Wang, Y., & Zhao, H. (2023). VectorMapNet: End-to-end vectorized HD map learning. Proceedings of the 40th International Conference on Machine Learning, Honolulu, Hawaii, USA.

Lu, Y., Wang, W., Bai, R., Zhou, S., Garg, L., Bashir, A. K., Jiang, W., & Hu, X. (2025). Hyper-relational interaction modeling in multi-modal trajectory prediction for intelligent connected vehicles in smart cites. Information Fusion, 114, 102682. http://doi.org/10.1016/j.inffus.2024.102682

Madjid, N. A., Ahmad, A., Mebrahtu, M., Babaa, Y., Nasser, A., Malik, S., Hassan, B., Werghi, N., Dias, J., & Khonji, M. (2026). Trajectory prediction for autonomous driving: Progress, limitations, and future directions. Information Fusion, 126, 103588. http://doi.org/10.1016/j.inffus.2025.103588

Mohamed, A., Qian, K., Elhoseiny, M., & Claudel, C. (2020). Social-STGCNN: A social spatio-temporal graph convolutional neural network for human trajectory prediction. 2020 IEEE/CVF Conference on Computer Vision and Pattern Recognition (CVPR), Seattle, WA, USA. http://doi.org/10.1109/CVPR42600.2020.01443

Mohamed, A., Zhu, D., Vu, W., Elhoseiny, M., & Claudel, C. (2022). Social-Implicit: Rethinking trajectory prediction evaluation and the effectiveness of implicit maximum likelihood estimation. European Conference on Computer Vision, Cham: Springer Nature Switzerland. http://doi.org/10.1007/978-3-031-20047-2_27

Peng, Y., Zhang, G., Shi, J., Xu, B., & Zheng, L. (2022). SRAI-LSTM: A social relation attention-based interaction-aware LSTM for human trajectory prediction. Neurocomputing, 490, 258-268. http://doi.org/10.1016/j.neucom.2021.11.089

Sang, H., Chen, W., & Zhao, Z. (2026). Review of pedestrian trajectory prediction based on graph neural networks. Information Fusion, 127, 103727. http://doi.org/10.1016/j.inffus.2025.103727

Shi, L., Wang, L., Zhou, S., & Hua, G. (2023). Trajectory unified Transformer for pedestrian trajectory prediction. 2023 IEEE/CVF International Conference on Computer Vision (ICCV), Paris, France. http://doi.org/10.1109/ICCV51070.2023.00887

Si, Z., Ye, P., Xiong, G., Lv, Y., & Zhu, F. (2024). DATraj: A dynamic graph attention based model for social-aware pedestrian trajectory prediction. 2024 IEEE Intelligent Vehicles Symposium (IV), Jeju Island, Korea. http://doi.org/10.1109/IV55156.2024.10588693

Tang, Y., He, H., & Wang, Y. (2024). Hierarchical vector transformer vehicle trajectories prediction with diffusion convolutional neural networks. Neurocomputing, 580, 127526. http://doi.org/10.1016/j.neucom.2024.127526

Vaswani, A., Shazeer, N., Parmar, N., Uszkoreit, J., Jones, L., Gomez, A. N., Kaiser, L. U., & Polosukhin, I. (2017). Attention is All you Need. Advances in Neural Information Processing Systems 30 (NIPS 2017). http://doi.org/10.48550/arXiv.1706.03762

Wang, Z., Sun, Z., Luettin, J., & Halilaj, L. (2024). SocialFormer: Social interaction modeling with edge-enhanced heterogeneous graph Transformers for trajectory prediction. arXiv preprint arXiv:2405.03809 http://doi.org/10.48550/arXiv.2405.03809

Wang, Z., Zhang, J., Chen, J., & Zhang, H. (2024). Spatio-temporal context graph transformer design for map-free multi-agent trajectory prediction. IEEE Transactions on Intelligent Vehicles, 9(1), 1369-1381. http://doi.org/10.1109/TIV.2023.3329885

Wen, J., Lyu, N., & Liu, C. (2024). Safety evaluation and influential factor analysis of urban expressway diversion areas based on non-stationary conflict extremes. Transportation Research Record: Journal of the Transportation Research Board, 2678(12), 1813-1825. http://doi.org/10.1177/03611981241252836

Wen, J., Lyu, N., & Zheng, L. (2025). Exploring safety effects on urban expressway diverging areas: crash risk estimation considering extreme conflict types. International Journal of Injury Control and Safety Promotion, 32(1), 25-39. http://doi.org/10.1080/17457300.2024.2440940

Wen, J., Yamamoto, T., & Lyu, N. (2024). A two-stage behavioral model considering vehicle motion fluctuations for decision-making during lane changes in diverging areas. International Journal of Intelligent Transportation Systems Research, 22(3), 561-578. http://doi.org/10.1007/s13177-024-00416-1

Wu, Y., Chen, G., Li, Z., Zhang, L., Xiong, L., Liu, Z., & Knoll, A. (2021). HSTA: A hierarchical spatio-temporal attention model for trajectory prediction. IEEE Transactions on Vehicular Technology, 70(11), 11295-11307. http://doi.org/10.1109/TVT.2021.3115018

Xiang, J., Nan, Z., Song, Z., Huang, J., & Li, L. (2024). Map-free trajectory prediction in traffic with multi-level spatial-temporal modeling. IEEE Transactions on Intelligent Vehicles, 9(2), 3258-3270. http://doi.org/10.1109/TIV.2023.3342430

Xu, C., Li, M., Ni, Z., Zhang, Y., & Chen, S. (2022). GroupNet: Multiscale hypergraph neural networks for trajectory prediction with relational reasoning. 2022 IEEE/CVF Conference on Computer Vision and Pattern Recognition (CVPR), New Orleans, LA, USA. http://doi.org/10.1109/CVPR52688.2022.00639

Xu, P., Hayet, J., & Karamouzas, I. (2023). Context-aware timewise VAEs for real-time vehicle trajectory prediction. IEEE Robotics and Automation Letters, 8(9), 5440-5447. http://doi.org/10.1109/LRA.2023.3295990

Yang, Z., Wan, Y., Du, L., Zhang, W., Yang, X., & Han, Y. (2024). Vehicle trajectory prediction based on attention optimized with real-scene sampling. Systems science & control engineering, 12(1) http://doi.org/10.1080/21642583.2024.2347889

Yang, Z., Yang, J., Zhou, Y., Xu, Q., & Ou, M. (2025). Multimodal trajectory prediction for intelligent connected vehicles in complex road scenarios based on causal reasoning and driving cognition characteristics. Scientific Reports, 15(1), 7259. http://doi.org/10.1038/s41598-025-91818-y

Zhang, X., Liu, G., Liu, Z., Xu, N., Liu, Y., & Zhao, J. (2024). Enhancing vectorized map perception with historical rasterized maps. European Conference on Computer Vision, Cham. http://doi.org/10.1007/978-3-031-72643-9_25

Zhao, L., Zhou, W., Xu, S., Chen, Y., & Wang, C. (2025). Multi-agent trajectory prediction at unsignalized intersections: An improved generative adversarial network accounting for collision avoidance behaviors. Transportation Research Part C: Emerging Technologies, 171, 104974. http://doi.org/https://doi.org/10.1016/j.trc.2024.104974

Zheng, F., Wang, L., Zhou, S., Tang, W., Niu, Z., Zheng, N., & Hua, G. (2021). Unlimited neighborhood interaction for heterogeneous trajectory prediction. 2021 IEEE/CVF International Conference on Computer Vision (ICCV), Montreal, QC, Canada. http://doi.org/10.1109/ICCV48922.2021.01292